\documentclass{ecai} 

\usepackage{latexsym}
\usepackage{hyperref}
\usepackage{amssymb}
\usepackage{amsmath}
\usepackage{amsthm}
\usepackage{booktabs}
\usepackage{enumitem}
\usepackage{graphicx}
\usepackage{xcolor}
\usepackage{color, colortbl}
\usepackage{subfigure}
\usepackage{wrapfig}
\usepackage{tabularx}

\def\eqref#1{equation~\ref{#1}}

\def\1{\bm{1}}

\def\vtheta{{\bm{\theta}}}
\def\vphi{{\bm{\phi}}}

\DeclareMathAlphabet{\mathsfit}{\encodingdefault}{\sfdefault}{m}{sl}
\SetMathAlphabet{\mathsfit}{bold}{\encodingdefault}{\sfdefault}{bx}{n}

\def\gA{{\mathcal{A}}}

\def\gD{{\mathcal{D}}}

\def\gM{{\mathcal{M}}}

\def\gS{{\mathcal{S}}}

\usepackage{amsmath}
\usepackage{bm}
\usepackage{bbm}
\usepackage{booktabs}
\usepackage{bbding}
\usepackage{amsmath}
\usepackage{multirow}
\usepackage{mathtools}
\usepackage{enumitem}
\usepackage[hang,flushmargin]{footmisc}
\definecolor{Gray}{gray}{0.9}
\definecolor{myred}{HTML}{B51800}
\definecolor{mygreen}{HTML}{017100}
\setitemize{itemsep=-2pt,leftmargin=*}
\setenumerate{itemsep=-2pt,leftmargin=*}

\newcommand{\Dpre}{\gD_{\mathrm{pre}}}
\newcommand{\Dfine}{\gD_{\mathrm{fine}}}
\newcommand{\Dnon}{\gD_{\mathrm{non}}}

\newcommand{\Spre}{\gS_{\mathrm{pre}}}
\newcommand{\Snon}{\gS_{\mathrm{non}}}

\newcommand{\Npre}{N_{\mathrm{pre}}}
\newcommand{\Nfine}{N_{\mathrm{fine}}}

\newcommand{\BibTeX}{B\kern-.05em{\sc i\kern-.025em b}\kern-.08em\TeX}
\renewcommand*{\sectionautorefname}{Section}

\begin{document}

\begin{frontmatter}
\paperid{2049} 

\title{Inside the Black Box: Detecting Data Leakage in Pre-trained Language Encoders}

\author[a]{\fnms{Yuan}~\snm{Xin}}
\author[a]{\fnms{Zheng}~\snm{Li}}
\author[b]{\fnms{Ning}~\snm{Yu}}
\author[a]{\fnms{Dingfan}~\snm{Chen}\thanks{Corresponding Author. Email: dingfan.chen@outlook.com}}
\author[a]{\fnms{Mario}~\snm{Fritz}}
\author[a]{\fnms{Michael}~\snm{Backes}}
\author[a]{\fnms{Yang}~\snm{Zhang}}

\address[a]{CISPA Helmholtz Center for Information Security}
\address[b]{Netflix Eyeline Studios}

\begin{abstract}
Despite being prevalent in the general field of Natural Language Processing (NLP), pre-trained language models inherently carry privacy and copyright concerns due to their nature of training on large-scale web-scraped data. In this paper, we pioneer a systematic exploration of such risks associated with pre-trained language encoders, specifically focusing on the membership leakage of pre-training data exposed through downstream models adapted from pre-trained language encoders—an aspect largely overlooked in existing literature. Our study encompasses comprehensive experiments across four types of pre-trained encoder architectures, three representative downstream tasks, and five benchmark datasets. Intriguingly, our evaluations reveal, for the first time, the existence of membership leakage even when only the black-box output of the downstream model is exposed, highlighting a privacy risk far greater than previously assumed. Alongside, we present 
in-depth analysis and insights toward guiding future researchers and practitioners in addressing the privacy considerations in developing pre-trained language models.
\end{abstract}

\end{frontmatter}

\section{Introduction}

Pre-trained language encoders (PLEs), exemplified by BERT~\citep{DCLT19}, underpin the recent advancements in the general field of natural language processing and have found widespread use across various application scenarios~\citep{WDSCDMCRLFB19,VSPUJGKP17,MSS19}. Commonly, PLEs are trained on large-scale text corpora to encapsulate general linguistic patterns, subsequently fine-tuned to refine their internal representations for specific downstream tasks~\citep{SQXH19,SJZLCSFYW21}.
Typically, model developers leverage PLEs by integrating a few layers, explicitly designed for the intended downstream tasks, to these pre-trained encoders, followed by fine-tuning the model on downstream data. This process enables the models to be customized for a broad range of tasks, such as text classification, named entity recognition (NER), and question answering (Q\&A).

Despite the extensive application of PLEs, the inherent risks associated with information leakage and copyright infringement of the pre-training data through the use of PLEs remain largely under-explored. Specifically, it remains unclear \textit{whether we can determine if a PLE-based language model, given a piece of text and black-box access, has been pre-trained on the provided text}. While this problem can be formulated as an instance of Membership Inference Attacks (MIAs),
existing MIAs typically necessitate assumptions about the attackers' access that may not align with practical usage scenarios involving PLEs~\cite{MGUBS22,JRY21,SIHS21}. In particular, all prior attack surfaces require direct access to target models that are perfectly trained for the data the attacker aim to infer, with no observed discrepancy. 

To fill this gap, we initiate the first systematic study of the information leakage risks inherent in PLEs by investigating the vulnerability of PLE frameworks  %
to adversaries attempting to infer their pre-training data.
Specifically, we introduce an attack pipeline for the most realistic scenario, where service providers build downstream models by integrating PLEs internally and only grant access to the downstream models (instead of direct access to the PLEs) in a black-box manner.

To systematically explore the potential risks associated with the usage of PLEs, we perform an extensive experimental investigation %
spanning four distinct PLE architectures and five downstream datasets across three representative downstream tasks.
Intriguingly, our evaluations uncover considerable data membership leakage within PLEs, even when only the output of the final downstream model is exposed. This leakage persists irrespective of the PLE architecture and the type of the downstream tasks, indicating a more severe risk than previously anticipated,  especially considering that this challenging setting has been ostensibly viewed as ``safe'' for usage.
Our study is further enriched by a comprehensive analysis that offers key insights into the primary factors associated with the privacy risk of PLEs, with which we aim to increase model developers' awareness of PLE vulnerabilities and to motivate the incorporation of privacy considerations into model design and training for privacy-preserving downstream usage.

\section{Related work}
\label{gen_inst}
\paragraph{Pre-trained Language Encoders (PLEs).} 
The use of PLEs is pervasive in the NLP domain due to their ability to capture generic linguistic characteristics of natural languages, which are universally beneficial for various downstream tasks that rely on the semantics of the representation~\citep{DCLT19, BMRSKDNSSAAHKHCRZWWHCSLGCCBMRSA20, LLGGMLSZ20}. In particular, PLEs stand out for their relatively lightweight usage and the flexibility they offer by providing semantic-aware embeddings, despite the recent development of large-scale decoder-based pre-trained models like GPT-4~\citep{brown2020language} and LAMMA~\citep{touvron2023llama}.
We focus on the most predominant instance of PLEs, specifically, BERT~\citep{DCLT19}, along with its prevalent variants~\citep{LCGGSS20, LOGDJCLLZS19, YDYCSL19}. 
 
\begin{figure*}[!t]
    \centering
    \includegraphics[width=\textwidth,trim={5 60 5 25}, clip]{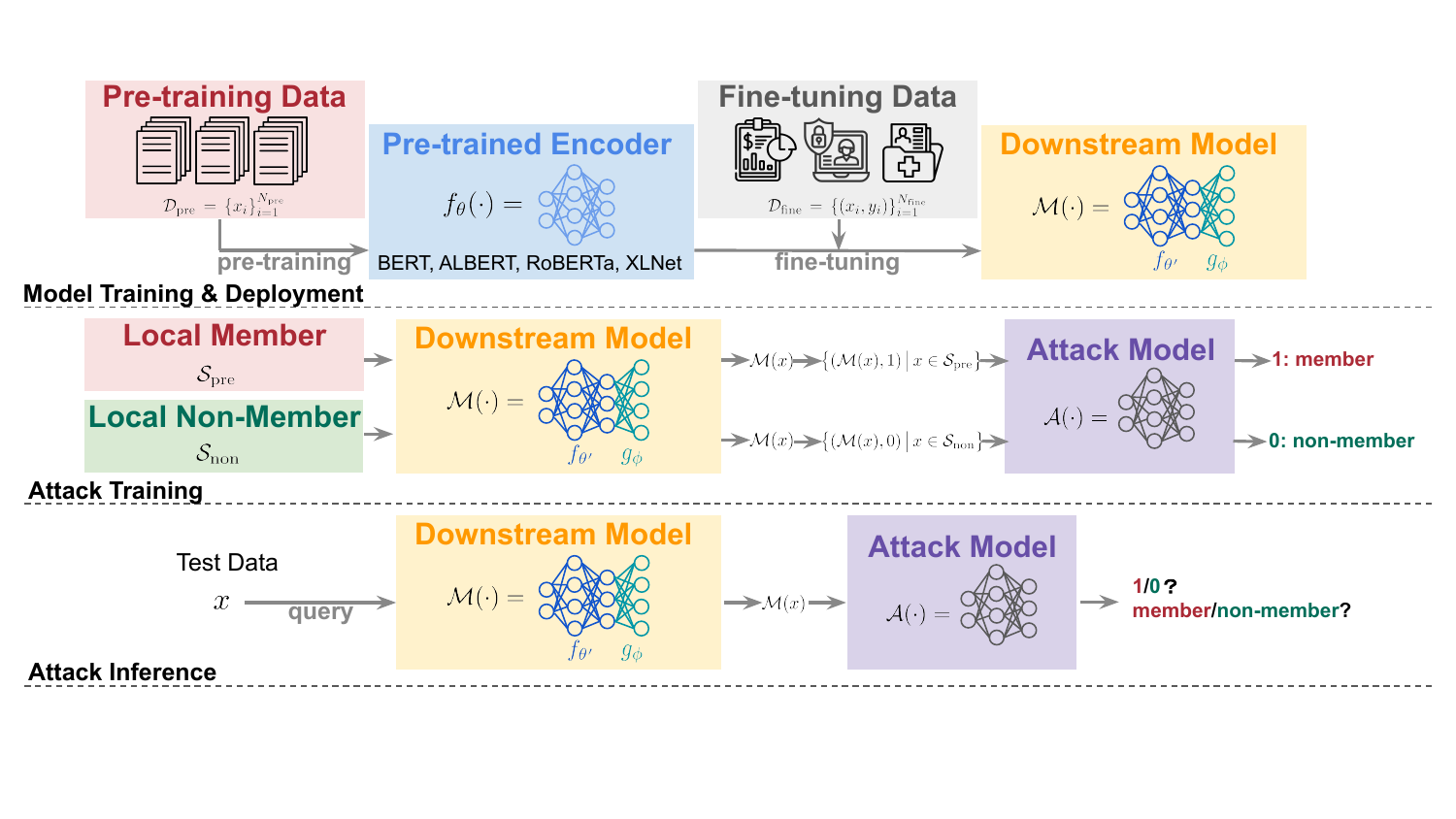}
    \caption{Overview of the workflow.}
    \label{fig:overview}
    \vspace{15pt}
\end{figure*}

\paragraph{Downstream Tasks.} 
PLEs are typically adopted within the \textit{pre-train and fine-tune} paradigm, where PLEs are pre-trained on large corpora through self-supervised learning (e.g., masked language modeling and next sentence prediction), and are subsequently integrated into downstream models while being fine-tuned to achieve specific objectives for various tasks~\cite{ramponi2020neural}. In \sectionautorefname~\ref{sec:method}, we introduce our attack pipeline, designed for seamless applicability to diverse downstream tasks. For our experimental evaluation, we focus on the representative tasks, namely text classification, named entity recognition (NER), and question \& answering (Q\&A).

\paragraph{Data Leakage of Pre-training Models.}
Pre-training on large-scale web-scraped data offers substantial advantages in learning generic linguistic representations, but may raise privacy and copyright concerns~\citep{LYKKKSK20}. This is particularly relevant in the context of stringent legal regulations, such as the General Data Protection Regulation (GDPR). 
Recent studies have highlighted these concerns, demonstrating privacy leakage of the training data for encoder-decoder models trained from scratch~\cite{SS19} and deployed decoder models~\cite{carlini2021extracting}, as well as of the fine-tuning data used in the pre-training fine-tuning pipeline~\cite{mireshghallah2022memorization}.
In particular, existing studies on pre-trained (encoder) models~\citep{MGUBS22,JRY21,tirumala2022memorization} assumes direct access of the attacker to the encoder. This, however,  may not be a practical presumption as PLEs are typically integrated internally into the downstream service models.

In our work, we delve into a more realistic and challenging scenario: our study does not rely on direct access to the original PLEs that have been pre-trained with the data we aim to infer. 
This more closely mirrors practical usage scenarios but implies a certain level of discrepancy between the target PLEs and the final output we have access to, which then raises the need for a more refined attack design to effectively extract private information. 
Moreover, this discrepancy is further amplified by the fine-tuning process, which adjusts the parameters of the PLEs to better suit the new downstream task. These inconsistencies complicate the assessment of a model's vulnerability to attacks, potentially leading to previous hasty claims regarding the ``safe'' usage of PLEs, while we question such claims with our comprehensive evaluation.

\section{Formulation}
\label{sec:formulation}
\subsection{Target Models}
We denote a PLE as $f_{\vtheta}$ (parametrized by $\vtheta$) which maps 
a given textual input $x$ to its embedding $f_{\vtheta}(x)$. PLE is trained on unlabelled \textit{pre-training dataset} $\Dpre$\,$=$\,$\{x_i\}_{i=1}^{\Npre}$ with self-supervision objectives. The downstream model $\gM$ is constructed by appending layers, which map the embeddings to the final output prediction space, atop the PLEs. Specifically, $\gM(x)$\,$=$\,$g_{\vphi}\big(f_{\vtheta}(x)\big)$, where $g_{\vphi}$ denotes the task-specific downstream layers with parameters $\vphi$. The downstream model
$\gM$ is fine-tuned on the labelled \textit{fine-tuning dataset} $\Dfine$\,$=$\,$\{(x_i, y_i)\}_{i=1}^{\Nfine}$ with $y_i$ denoting the task-specific labels, e.g., class index for text classification tasks. During the fine-tuning process, the parameters for both the encoder $\vtheta$ and the downstream layers $\vphi$ are updated. If necessary for clarity, we may denote the updated encoder parameters as $\vtheta'$ to distinguish them from the pre-trained model's parameters. The dataset size $\Nfine$ is typically much smaller than $\Npre$ due to the higher difficulty and workload involved in obtaining labelled data.

\subsection{Membership Inference Attacks} 
A membership inference attack refers to a privacy attack where an adversary attempts to determine whether a particular sample was part of the training set used to train a target machine learning model~\citep{HSSDYZ21,CCNSTT21,YMMS21}. In this context, all training data are considered as ``\textit{members}'' while any data not included in the training set (i.e., the \textit{unseen} data) are regarded as ``\textit{non-members}''.\\
We formalize the attacker as a binary classification model $\gA$, which receives a query sample $x$ and the corresponding output from a target model $\gM(x)$. In this work, the attacker's goal is to infer whether the given sample is inside the pre-training set of the target model, i.e., $\gA(x,\gM(x))$\,$=$\,$\mathbbm{1}\big[x$\,$\in$\,$\Dpre\big]$ with $\mathbbm{1}$ denoting the indicator function and $\Dpre$ representing the pre-training dataset.

\begin{figure*}[t]
\centering 
\subfigure[Pre-trained BERT]{
\begin{minipage}[b]{0.31\textwidth}
\label{fig:tsne_pretrained}
    \includegraphics[width=1.0\textwidth]{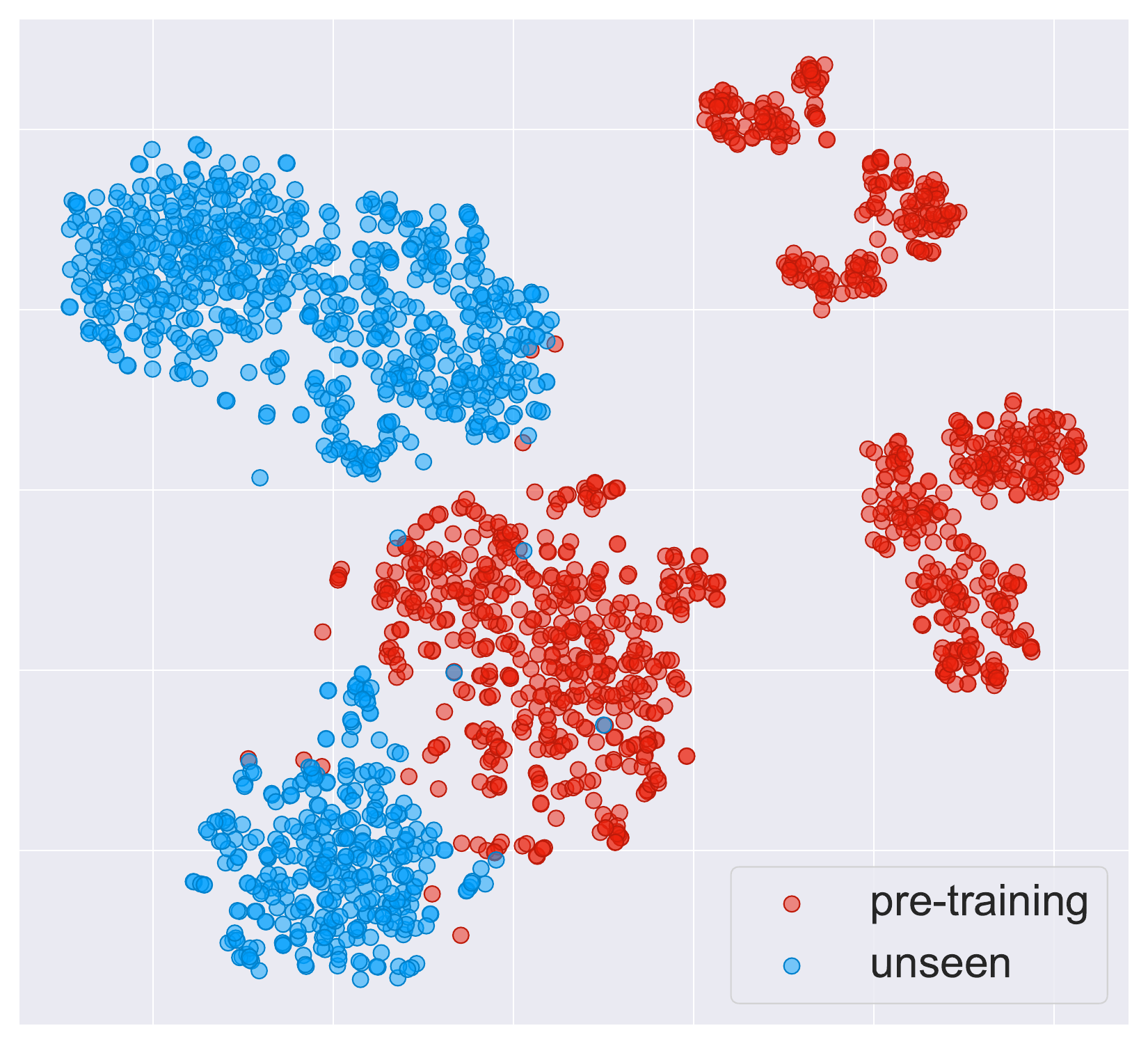}
\end{minipage}\hfill

}
\subfigure[Fine-tuned BERT]{
\begin{minipage}[b]{0.31\textwidth}
\label{fig:tsne_finetuned}
    \includegraphics[width=1.0\textwidth]{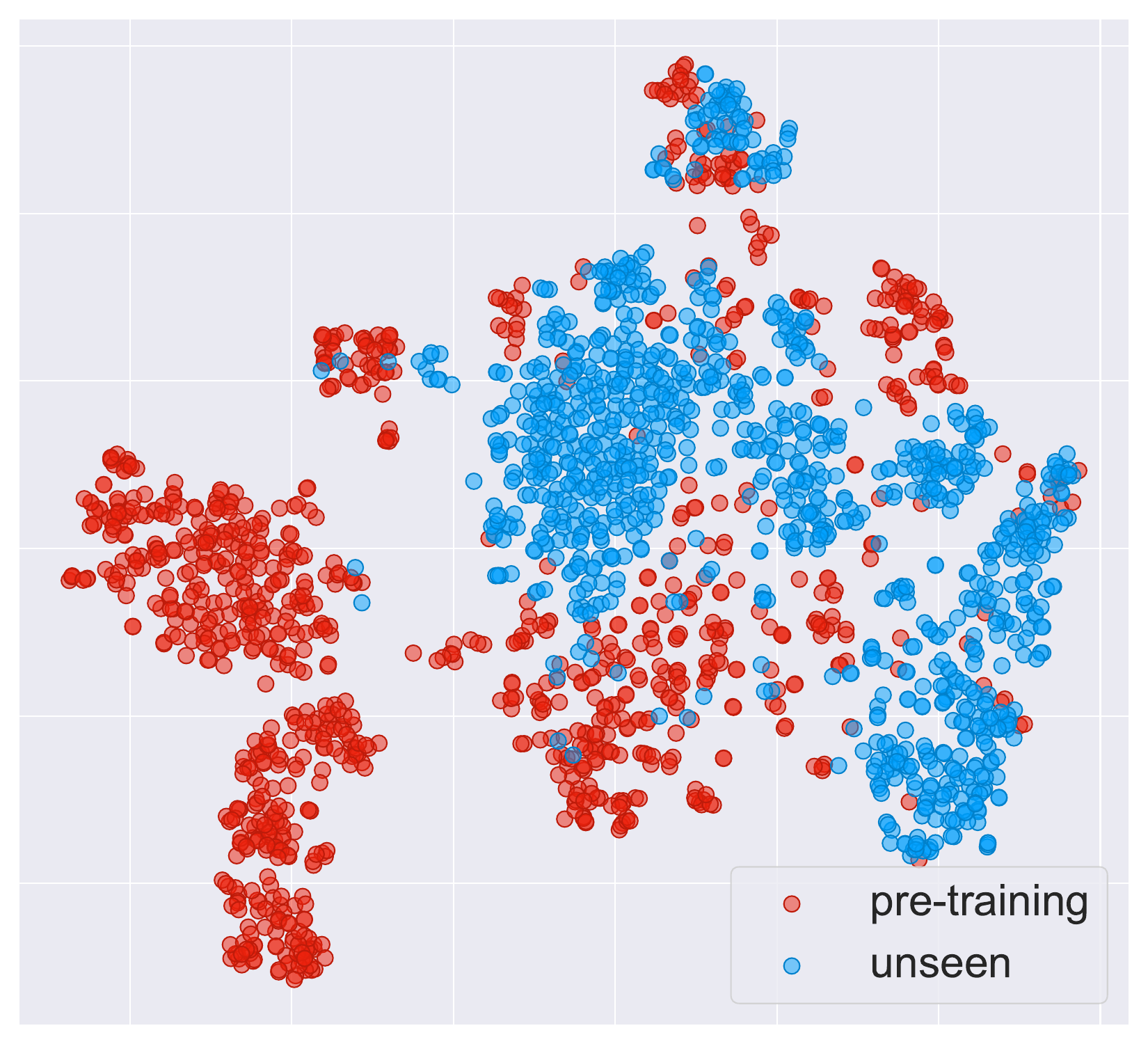}
\end{minipage}\hfill
}
\subfigure[Downstream model]{
\begin{minipage}[b]{0.31\textwidth}
\label{fig:tsne_downstream}
    \includegraphics[width=1.0\textwidth]{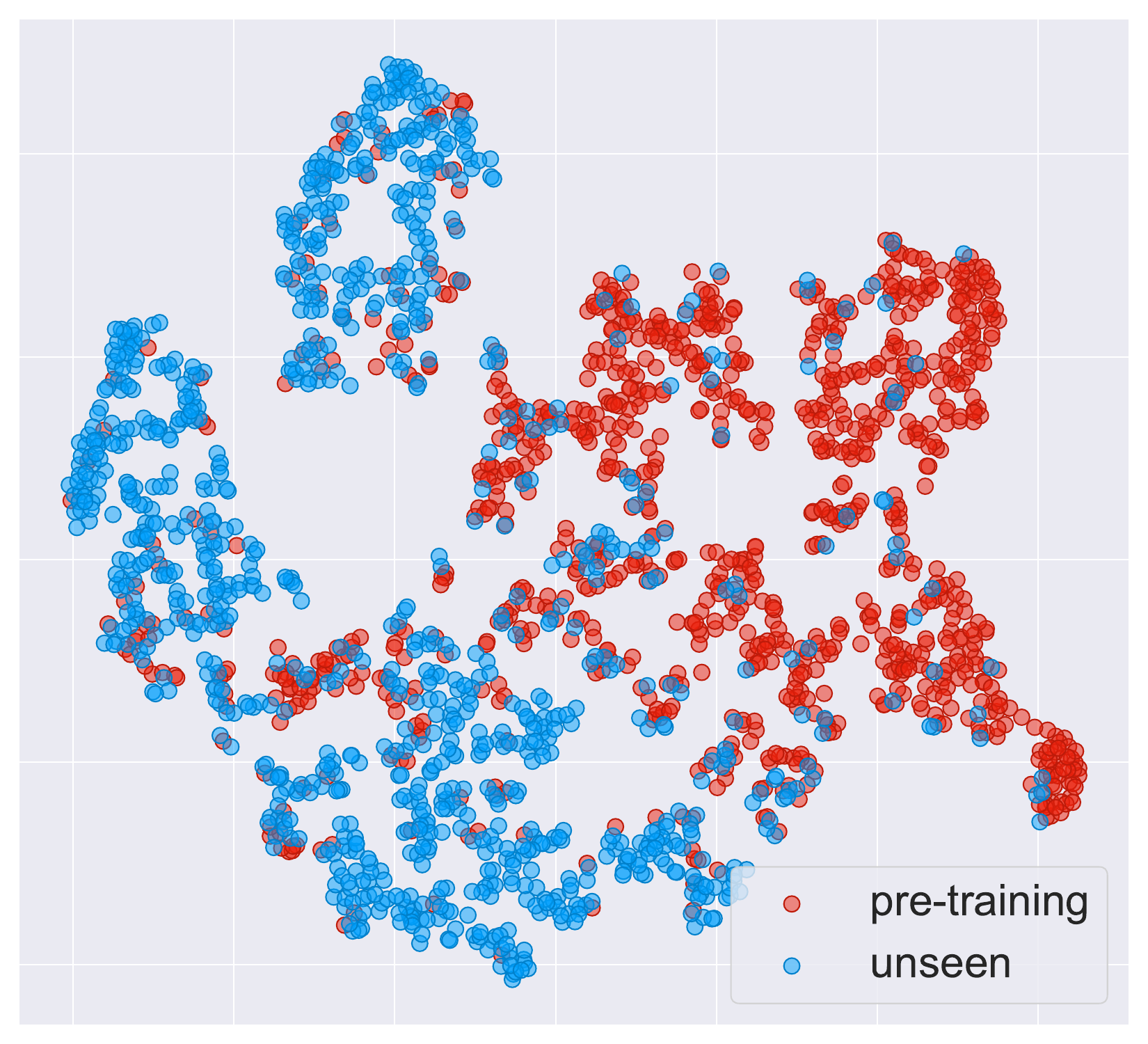}
\end{minipage}\hfill
}
\caption{t-SNE visualization of BERT embeddings. The \textit{pre-training} and \textit{unseen} samples are plotted as \textcolor[HTML]{BF1E0E}{red} and \textcolor[HTML]{1E92D6}{blue} dots, respectively. \textbf{(a)} Embeddings directly obtained from PLEs, i.e.,  $f_\vtheta(x)$. \textbf{(b)} Embeddings obtained from the encoder after fine-tuning, which corresponds to $f_{\vtheta'}(x)$ with ${\vtheta'}\neq \vtheta$. \textbf{(c)} Embeddings obtained from the downstream model, i.e., $g_\vphi(f_{\vtheta'}(x))$. Fine-tuning is conducted on the AG's News dataset. %
}
\label{fig:tsne}
\vspace{15pt}
\end{figure*}

\section{Attack Method}

\label{sec:method}
\subsection{Threat Model}
\label{subsec:threat_model}
\textbf{Attacker's Goal.}
In this work, we focus on the privacy leakages of the \textit{pre-training} data of PLEs. %
Specifically, the attacker aims to infer whether a given sample was part of the pre-training dataset used to train the PLE. The attacker has access to the downstream model, denoted as $\gM(\cdot) = g_\vphi\big(f_{\vtheta'}(\cdot)\big)$, where $f_{\vtheta'}$ represents the fine-tuned PLE, and $g_\vphi$ is the downstream task-specific head.
The key challenge for the attacker lies in the fact that the PLE may have undergone fine-tuning, meaning its parameters have been updated from their original pre-trained state, i.e., $\vtheta' \neq \vtheta$. Despite this, the attacker seeks to infer the membership status of the pre-training data that influenced the initial training of the PLE, even after fine-tuning has altered its parameters.

\paragraph{Attacker's Background Knowledge}
Typically, the background knowledge considered for MIAs falls into two dimensions: (1) the architecture of target models; (2) the distribution of target pre-training dataset. 

Along the first dimension, we consider the most general and realistic setting where (1) the attacker has no knowledge of the architecture of the target PLEs; (2) the attacker has only \textit{black-box} access to the downstream model $\gM$, which implies it can only input queries and receive predictions without knowing the internals; (3) the accessible downstream model $\gM$ has been fine-tuned to adapt to downstream task, resulting in parameter updates of the PLEs and a potential information loss regarding the membership of its pre-training data. This scenario closely mirrors real-world usage of PLEs,  as service providers typically share task-specific models adapted from PLEs to the public as part of ``machine learning as a service'', leading the attackers to access the output of the downstream model $\gM$ rather than direct access to the target PLEs. Notably, such scenario is largely under-explored in the general field of trustworthy learning, potentially giving rise to a false sense of privacy preservation within this use case scenario.

Regarding the second dimension, in line with previous studies~\cite{SZHBFB19}, we presume that the attacker has access to a very small subset of the pre-training data, denoted as $\Spre$ (i.e., $\Spre \subset \Dpre$ and $|\Spre|\ll|\Dpre|$). The attacker also has the ability to compile a small set of local non-member data, denoted as $\Snon$. These datasets are subsequently used to train the attack model $\gA$ (refer to \sectionautorefname~\ref{subsec:attack_pipeline}) while we detail the investigation (and potential relaxation) of the construction of such dataset in \sectionautorefname~\ref{total experiment}. Such an assumption may not be implausible in real-world scenarios, considering that PLEs typically utilize billions of web-scraped data samples for pre-training. It is conceivable that an attacker might manage to gather a very small fraction of such voluminous pre-training data. 
Moreover, this assumption is more feasible than the daunting task of collecting billions of local samples to construct a ``shadow model'' that is often adopted in previous studies for membership inference~\cite{SSSS17,SZHBFB19}.

\subsection{Intuition}
\label{sec:intuition}
The main underlying principle of MIAs is the strategic use of the memorization effect of member data in target models~\citep{SSSS17,SZHBFB19,HLXCZ22,LLHYBZ222}. Modern deep learning models, while exhibiting substantial expressive capacity due to their large number of parameters, are also susceptible to inherent generalization issues. This is primarily attributed to the empirical risk minimization formulation, which tends to promote overlearning/overfitting and memorization of training data. As a result, models typically display distinct behaviors when queried with both member and non-member data, which can be exploited by potential attackers to differentiate between the target model's training and unseen data~\citep{CCNSTT21,ROF202,HLXCZ22,MGUBS22}.

\subsection{Attack Pipeline}
\label{subsec:attack_pipeline}

\textbf{Training.}
The construction of the attack training dataset is achieved through pairing the black-box output from the target downstream model $\gM$ when queried using the local data  (\sectionautorefname~\ref{subsec:threat_model}), with the binary membership indicators, where ``$1$'' represents known pre-training samples and ``$0$'' denotes unseen samples. These indicators act as the target prediction labels for the attacker.

Formally, the attack model $\gA$ (represent as a neural network) is trained on $\big\{(\gM(x),1)\,\big|\,x\in \Spre\big\}\mathrel{\cup} \big \{(\gM(x),0)\,\big|\,x\in \Snon \big\}$ with $\Spre$ and $\Snon$ being the local pre-training and unseen sample set (\sectionautorefname~\ref{subsec:threat_model}), respectively. Specifically, the attack model is given the model responses $\gM(x)$ as input, and trained with the binary cross entropy objective while taking the membership indicator as the target prediction labels. 

\paragraph{Inference.}
After being trained (on the local dataset) to distinguish systematic disparities within the model's responses (see \figureautorefname~\ref{fig:tsne} for a visual illustration) for its pre-training versus unseen data, the attacker is then able to process a query text input and determine its membership status.  Specifically, if the query input more closely aligns with the higher confidence score for the ``pre-training'' class as opposed to the ``unseen'' class, the attacker will predict it as a pre-training sample.
\begin{figure*}[t]
\centering 
\subfigure[Accuracy]{
    \includegraphics[width=0.32\textwidth]{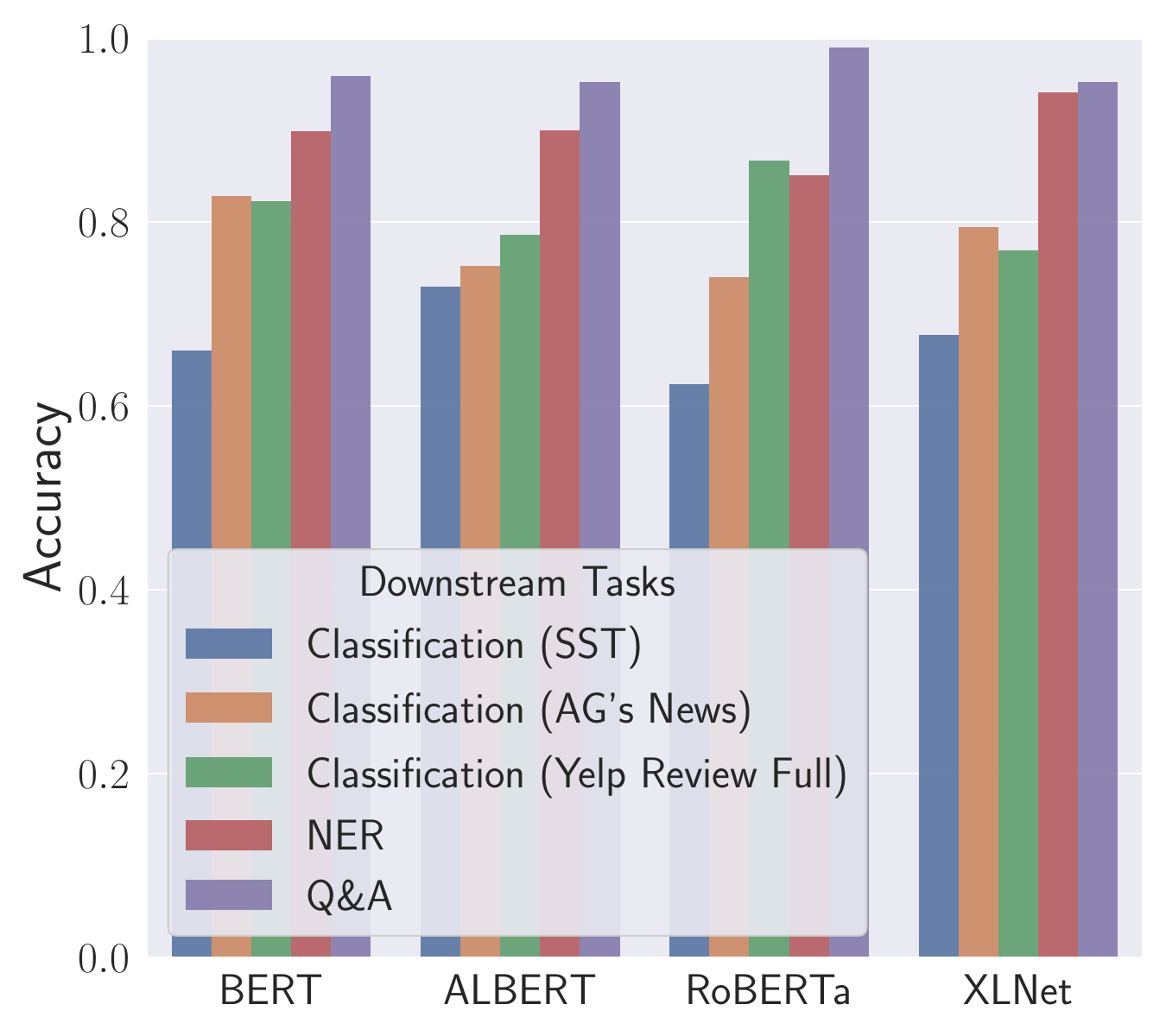}
}\hfill
\subfigure[Precision]{
    \includegraphics[width=0.32\textwidth]{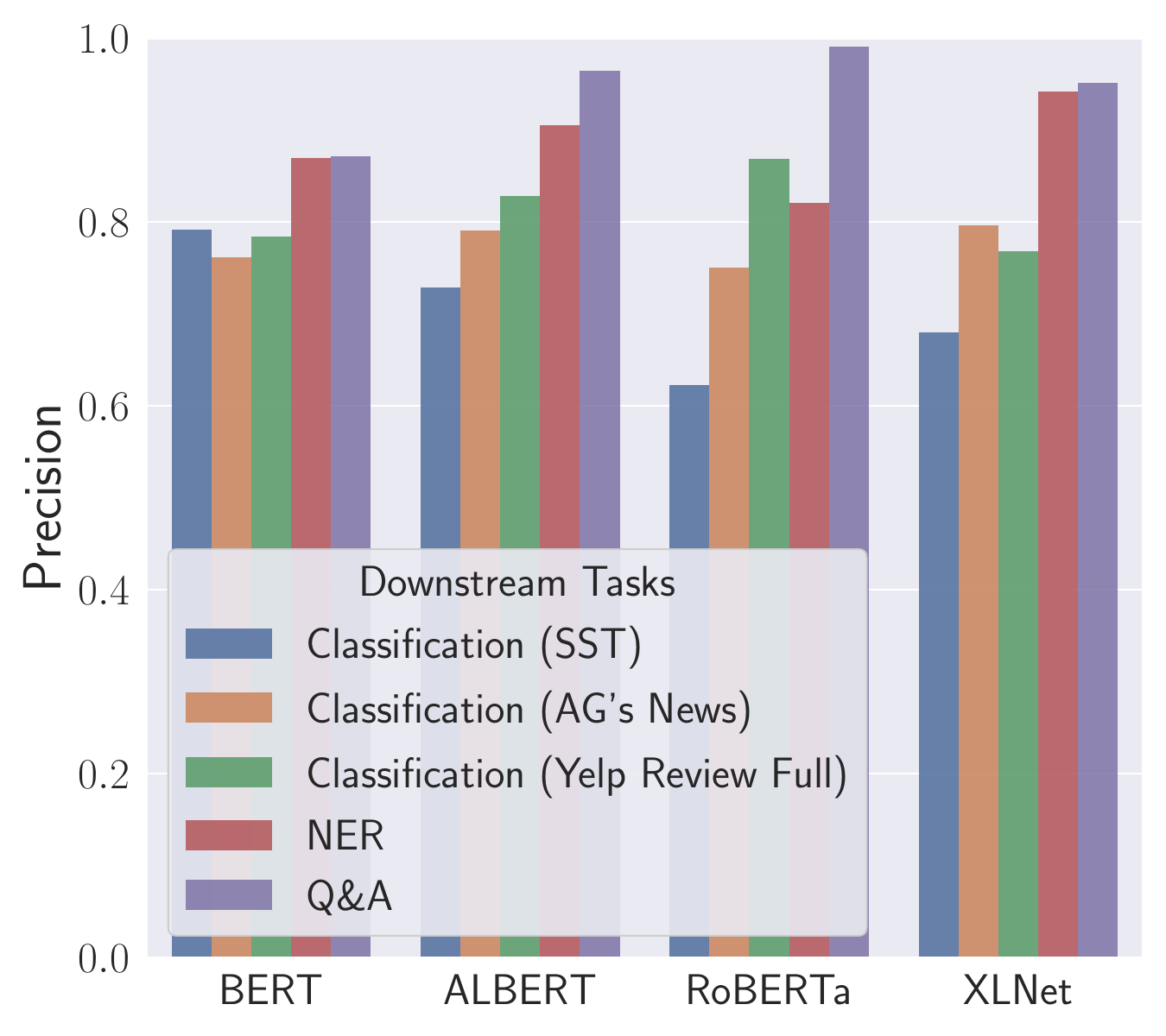}
}\hfill
\subfigure[Recall]{
    \includegraphics[width=0.32\textwidth]{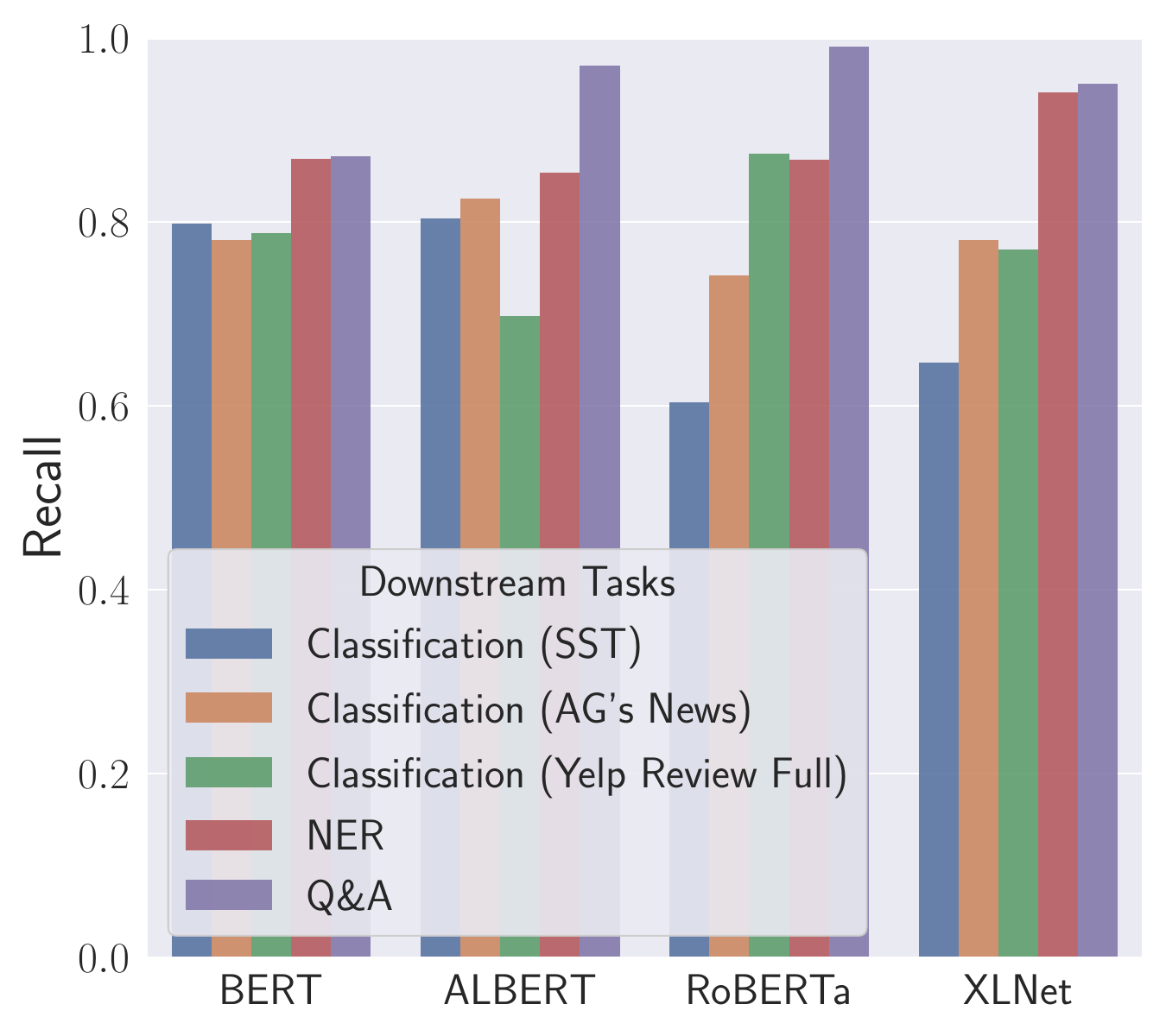}
}
\caption{Attack performance for different PLE architectures (BERT, ALBERT, RoBERTa, XLNet) on text classification, NER and Q\&A tasks. %
}
\label{fig:attack_performance}
\vspace{10pt}
\end{figure*}

 \begin{figure*}[!t]
	\centering

	\begin{minipage}[b]{0.49\textwidth}
		\subfigure[Relaxation-\MakeUppercase{\romannumeral1}]{
			\includegraphics[width=0.5\textwidth]{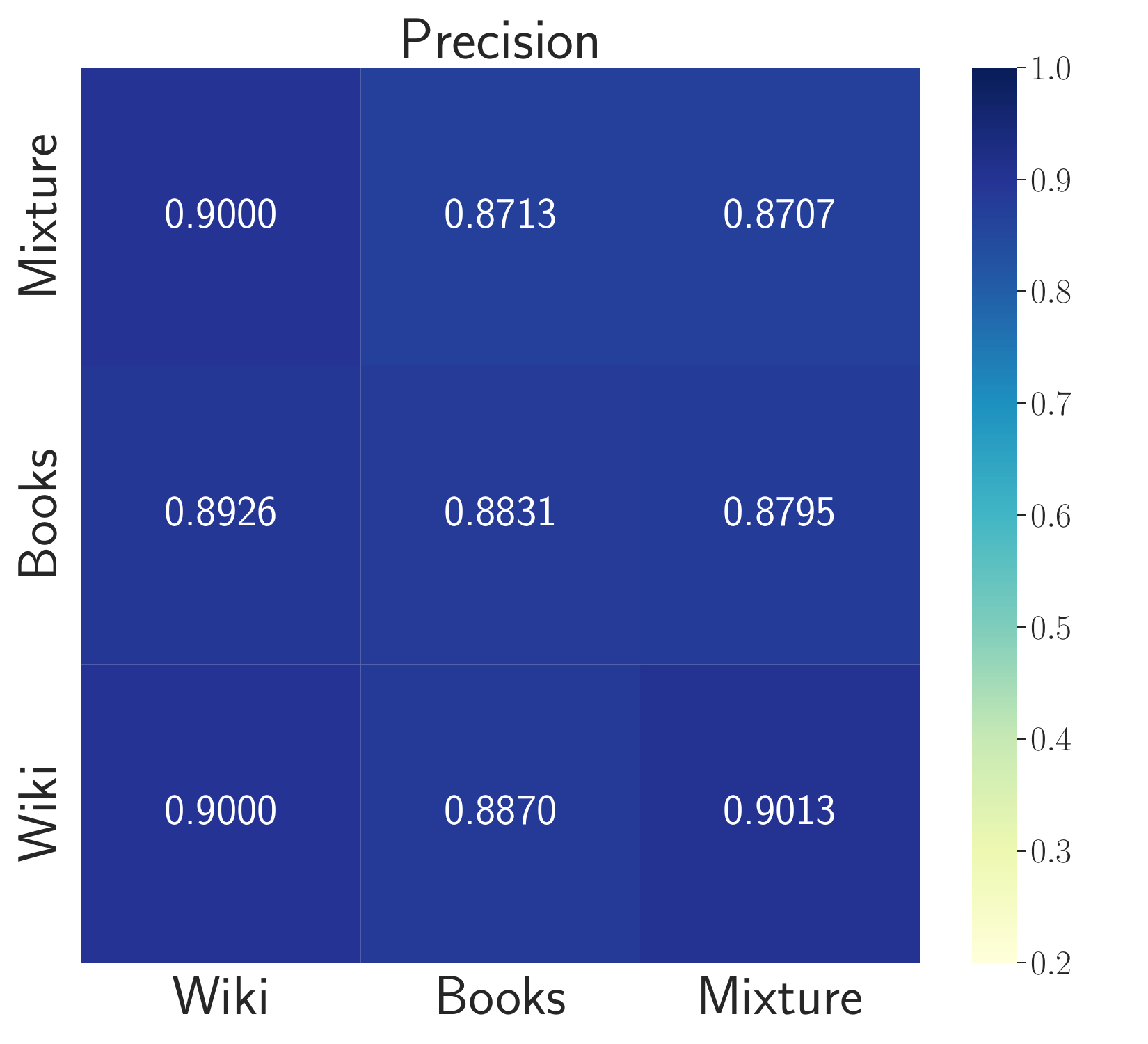} 
			\includegraphics[width=0.5\textwidth]{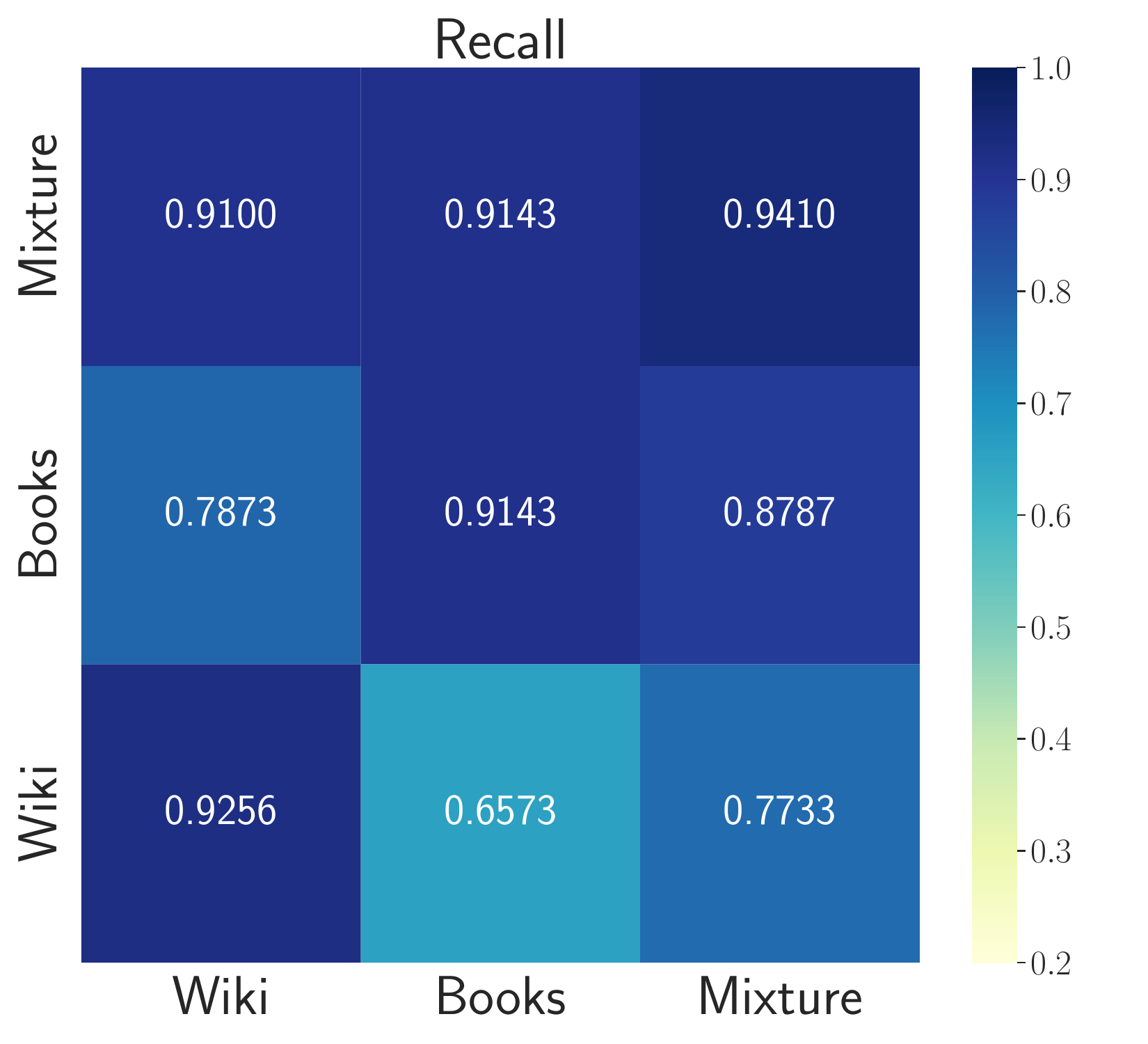}
			\label{heat_ner}
	}
	\end{minipage}
 	\begin{minipage}[b]{0.49\textwidth}
		\subfigure[Relaxation-\MakeUppercase{\romannumeral2}]{
			\includegraphics[width=0.5\textwidth]{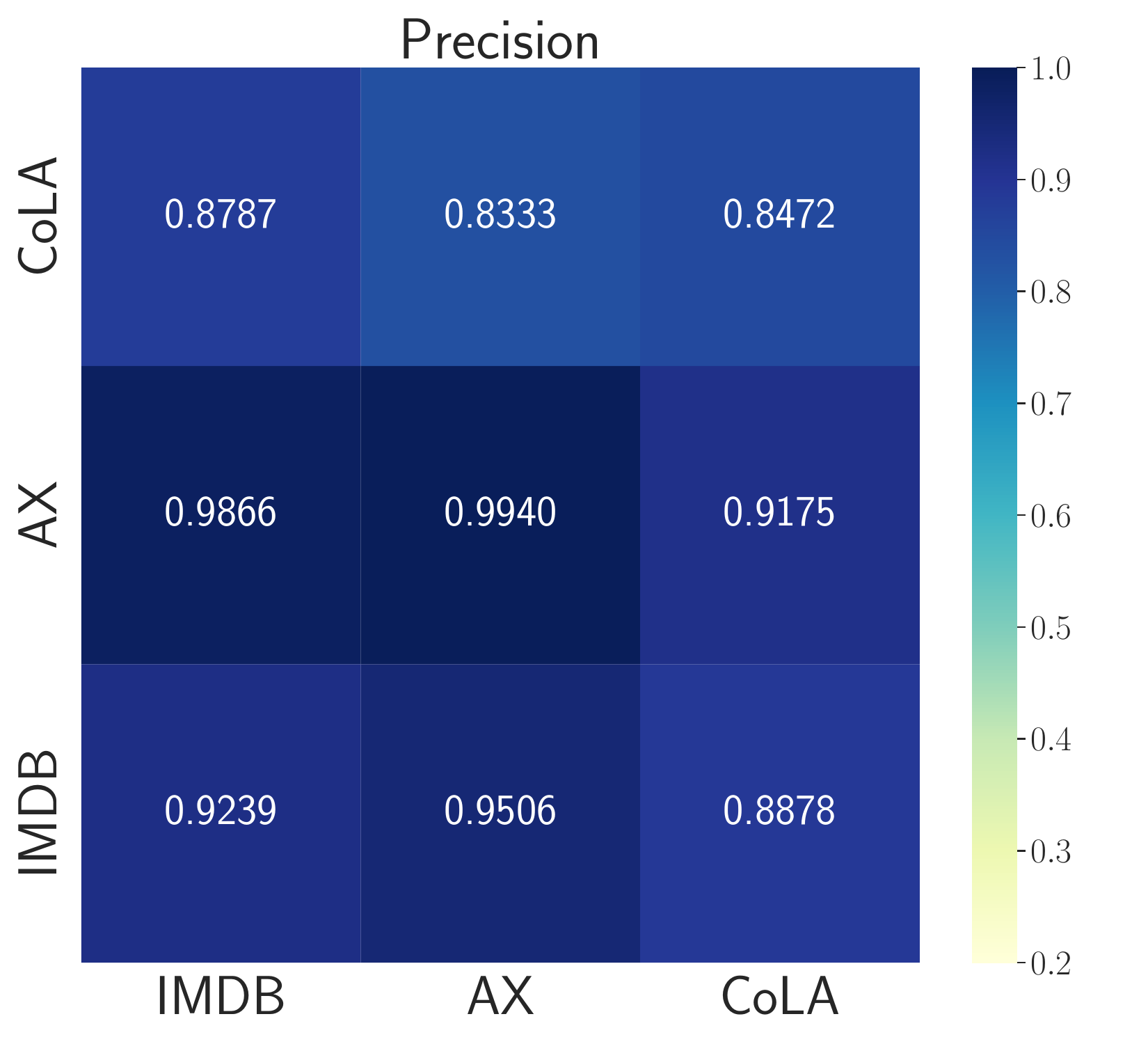} 
			\includegraphics[width=0.5\textwidth]{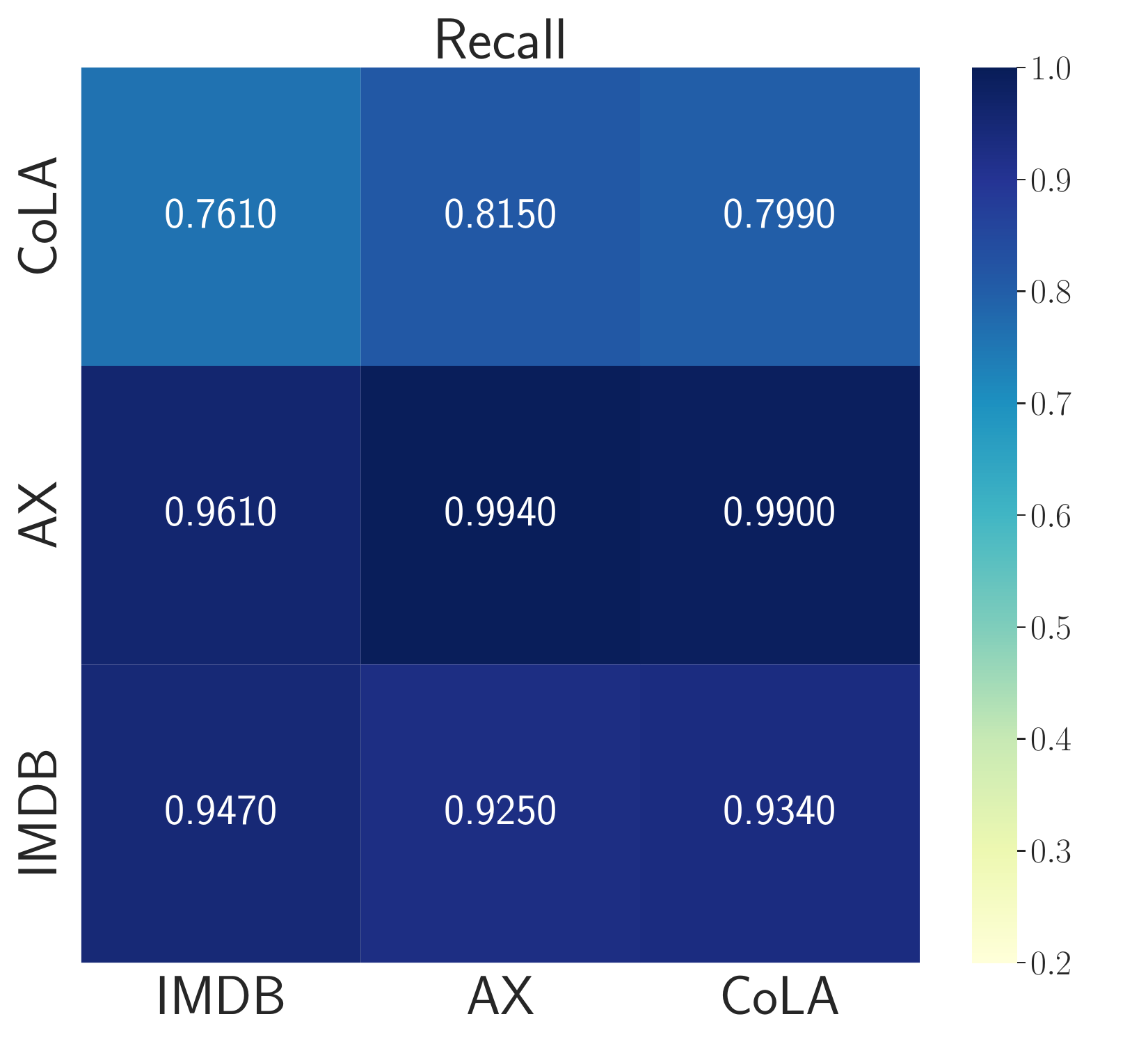}
			\label{heat_ner_nm}
	}
	\end{minipage}
	\caption{Attack performance with relaxation of \textit{pre-training} datasets (Relaxation-\MakeUppercase{\romannumeral1}) and relaxation of \textit{non-member} datasets (Relaxation-\MakeUppercase{\romannumeral2}) on NER downstream task: Wiki (Wikipedia), Books (BooksCorpus), Mixture (Wikepedia+BooksCorpus). X-axis: attack \textit{training} dataset. Y-axis: attack \textit{testing} dataset.  %
 }
	\label{fig:heatmap_member}
 \vspace{10pt}
\end{figure*}

\paragraph{Evaluation.}
The attack performance is evaluated on unseen testing samples that are distinct from the attack training set. 
While using arbitrary non-member testing data could be helpful for certain practical scenarios~\cite{maini2021dataset}, we opt for non-member testing samples that follows a similar distribution as the pre-training set, for a rigorous evaluation of MIA. 
Specifically, we adopt general-purpose natural texts from the GLUE benchmark as the non-member samples. We also explicitly test the setting in which we enforce the (semantic) similarity between the member and non-member instances. This is done by rephrasing the pre-training samples using a third-party language model (e.g., GPT-3.5) to generate the non-member samples for testing (See \sectionautorefname~\ref{sec:experiment:distribution} for detailed information).

\section{Experiments}
\subsection{Experimental Setup}
\label{experiment}
\paragraph{PLEs and Pre-training Data.}
In our experiments, we investigate four state-of-the-art architectures of PLEs: \textbf{BERT}~\citep{DCLT19}, \textbf{ALBERT}~\citep{LCGGSS20}, \textbf{RoBERTa}~\citep{LOGDJCLLZS19}, \textbf{XLNet}~\citep{YDYCSL19}.
We adopt well-trained PLEs, which are publicly available online, as the targets for our attacker\footnote{The PLEs used in this paper are downloaded from \url{https://huggingface.co/models}}. This is notably more realistic than the majority of existing work that is conducted in laboratory environments, which generally involves re-training the target model from scratch using a reduced set of training data samples. While different PLEs may employ their own pre-training data, we focus on two datasets—\textbf{Wikipedia} and \textbf{BooksCorpus}~\citep{DCLT19}—as these are commonly used across all four PLEs, and we categorize them as $\Dpre$ for the evaluation.

\paragraph{Downstream Models and Fine-tuning Data.}
As elaborated in \sectionautorefname~\ref{subsec:threat_model}, we evaluated the most challenging and realistic scenario in which the attacker can only access the outputs from downstream models that have been adapted from PLEs, while these downstream models may be utilized for any given tasks. For extensiveness, we considered six benchmark datasets for $\Dfine$, referring to three representative NLP topics. For \textit{text classification}, we adopt the  \textbf{SST}, \textbf{AG's News}, and \textbf{Yelp Review Full}~\citep{SPWCMNP13,ZZL15} datasets. The \textbf{CoNLL2003} dataset was used for the \textit{NER} studies~\citep{LFMHWL20}, while the \textbf{SQuADv1.0} dataset were chosen for the \textit{ Q\&A} task~\citep{BCW14}.
 
\paragraph{Attack Model and Attack Training\&Evaluation Data.}
For constructing the attacker's local training dataset, we randomly chose 30000 entries from the \textit{pre-training data} used across all PLEs, which represents a tiny fraction of the total dataset: the total size of Wikipedia and BooksCorpus are 16GB, (and we investigate the possibility to further reduce such fraction in \ref{fig:member_size}). We refer to this subset as the attacker's local members set, denoted as $\Spre$.
We then opted for third-party datasets to constitute the attacker's local non-member datasets, denoted as $\Snon$. These datasets are distinct from both the \textit{pre-training} and \textit{fine-tuning} datasets. More specifically, the local non-member datasets consist of 15,000 random samples from \textbf{IMDB}, \textbf{CoLA}, and \textbf{AX} datasets, which serve as part of the GLUE benchmark dataset~\citep{DCLT19}. For enhanced generalization, we use a mixture of these three datasets as non-members. 

For attack model, we constructed a three-layer Multilayer Perceptron (MLP) as the model architecture, which takes the output of downstream models $\gM(x)$ 
 as input and predict the binary membership indicator variable. Given the variety of downstream tasks, the dimension of the weight parameters in the first layer of the attack model is adjusted to suit different tasks.  
 The attack model is trained for 100 epochs with a learning rate of 1e-2, using the Adam~\citep{KB15} optimizer.
 By default, we set a 5:1 ratio for partitioning attack training and evaluation data, and adopt a balanced partition (maintaining a 1:1 ratio) for the member and non-member evaluation set. Additionally, we conduct a detailed investigation into the impact of the size and type of the attacker's local dataset on the attack performance in \figureautorefname~\ref{fig:heatmap_member}.
 
\paragraph{Evaluation Metrics.}
In line with previous studies~\citep{SSSS17,SZHBFB19,HLXCZ22}, we regard the MIA as a binary classification task and evaluate the attack performance across standard metrics including \textbf{accuracy}, \textbf{precision}, \textbf{recall}, and \textbf{F1-score}, while higher values of these metrics suggest a more effective attack and consequently a higher risk of data leakage from the PLEs.

\subsection{Experimental Results}
\label{total experiment}
\subsubsection{Attack Performance}
Firstly, we present the attack performance across standard metrics such as attack accuracy, precision, and recall in Figure~\ref{fig:attack_performance}. 
Our experimental results highlight that our attack generally demonstrates high performance, as evidenced by diverse metrics, across various PLE architectures and multiple downstream tasks. Within the context of MIA literature, this could be deemed a successful attack, given that it significantly surpasses the random guessing baseline of 0.5.
Taking the BERT model as an example, the attack executed across all different downstream models consistently demonstrates a high degree of effectiveness: The \textit{precision} for the classification, NER, and Q\&A tasks are around 0.77, 0.87, and 0.87 respectively. Meanwhile, the \textit{recall} for these tasks are around 0.79, 0.86, and 0.87. 
Notably, for all the investigated attack performance metrics, a value exceeding 0.6 is generally considered effective, whereas a value surpassing 0.8 is deemed significant.

On one hand, such findings suggest that membership leakage in PLEs is indeed a prevalent issue, irrespective of the type of downstream task, presenting considerably more severe potential privacy risks of PLEs than previously believed. On the other hand, such results may have broader implications in real-world scenarios for tasks such as privacy risk auditing and copyright authentication. For instance, our findings suggest the feasibility of leveraging our attack pipeline to detect potential data misuse during pre-training, requiring only black-box access to the final commercial models.

\subsubsection{Embedding Visualization}
While the quantitative results above demonstrate the vulnerability of PLEs to membership leakage, we delve deeper into the underlying reasons by visualizing the embeddings, which reveals potential systematic disparities between members and non-members that can be exploited by an attacker. As illustrated in \figureautorefname~\ref{fig:tsne}, we analyze PLEs' behavior in response to member and non-member samples for different scenarios within the pre-training fine-tuning framework.

Firstly, we directly feed 1000 pre-training and 1000 unseen samples into the original pre-trained BERT and embed the model output (i.e., $f_\vtheta(x)$) into a 2D space using t-Distributed Stochastic Neighbor Embeddings (t-SNE)\footnote{\url{https://github.com/DmitryUlyanov/Multicore-TSNE}} as seen in \figureautorefname~\ref{fig:tsne_pretrained}. Following this, we integrate the pre-trained BERT into a downstream model and conduct fine-tuning on AG's News datasets. Similarly, we plot the t-SNE visualizations of the fine-tuned BERT embedding (i.e., $f_{\vtheta'}(x)$), and the downstream model outputs (i.e., $g_{\vphi}(f_{\vtheta'}(x))$) in \figureautorefname~\ref{fig:tsne_finetuned} and \ref{fig:tsne_downstream}, respectively.

Our findings lead to several key observations together with insights that can be beneficial for future development of pre-trained models:
\begin{itemize}[topsep=2pt, partopsep=2pt,itemsep=2pt]
    \item  Original PLEs display a notable difference when being queried with its pre-training and unseen data samples, inevitably leaving cues for attackers to infer the membership information of the pre-training data samples. This necessitates privacy considerations and careful censorship of the pre-training dataset before PLEs are made publicly available.
    
    \item Intriguingly, PLEs consistently show a disparity in their response to the pre-training and unseen data, even post fine-tuning. This indicates the feasibility of an attack and the need for consider such vulnerability under practical usage scenarios of PLEs. 
    Moreover, this disparity remains evident in the outputs of downstream models, underscoring the plausibility of our proposed attack scenario.
    
    \item After fine-tuning on $\Dfine$ (which is disjoint from both the pre-training and non-member set in the evaluation), the difference between the target model‘s responses on pre-training and unseen data tends to diminish. This trend could be attributed to the model forgetting effect~\cite{goodfellow2013empirical}. However, a certain level of disparity still remains, as the general representation learned on the pre-training dataset may retain its utility for downstream tasks and thus be preserved during fine-tuning. 
    
    \item 
   In line with the data processing inequality principle—that is, the membership information is fully contained in the PLEs and any additional downstream layers will only decrease the available information—the final outputs from the downstream models exhibit less divergence between samples from $\Dpre$ and $\Dnon$ than the intermediate responses provided by the PLEs. This observation suggests a more challenging (yet realistic) scenario considered in our study, in comparison to previous studies where the fine-tuned encoder is directly accessible by potential attackers.
\end{itemize}

\vspace{8pt}
In summary, our visualizations qualitatively show that membership leakage in PLEs' pre-trained data persists, perhaps surprisingly, even when such PLEs have undergone the fine-tuning process and only indirect access through the black-box output of downstream models is available to potential adversaries.

 \begin{figure*}[!t]
\centering 
\subfigure[Accuracy]{
    \includegraphics[width=0.32\textwidth]{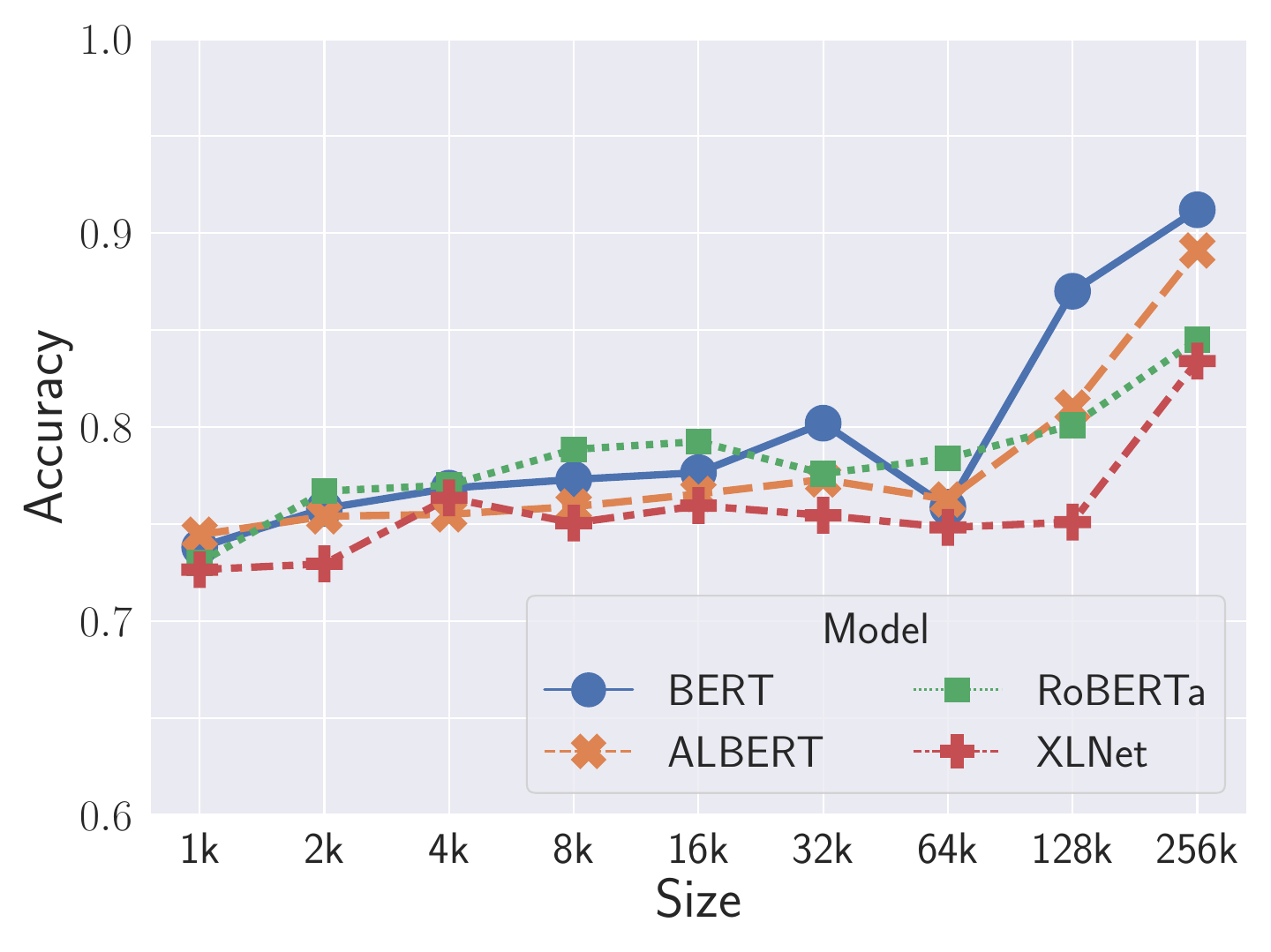}
}\hfill
\subfigure[Precision]{
    \includegraphics[width=0.32\textwidth]{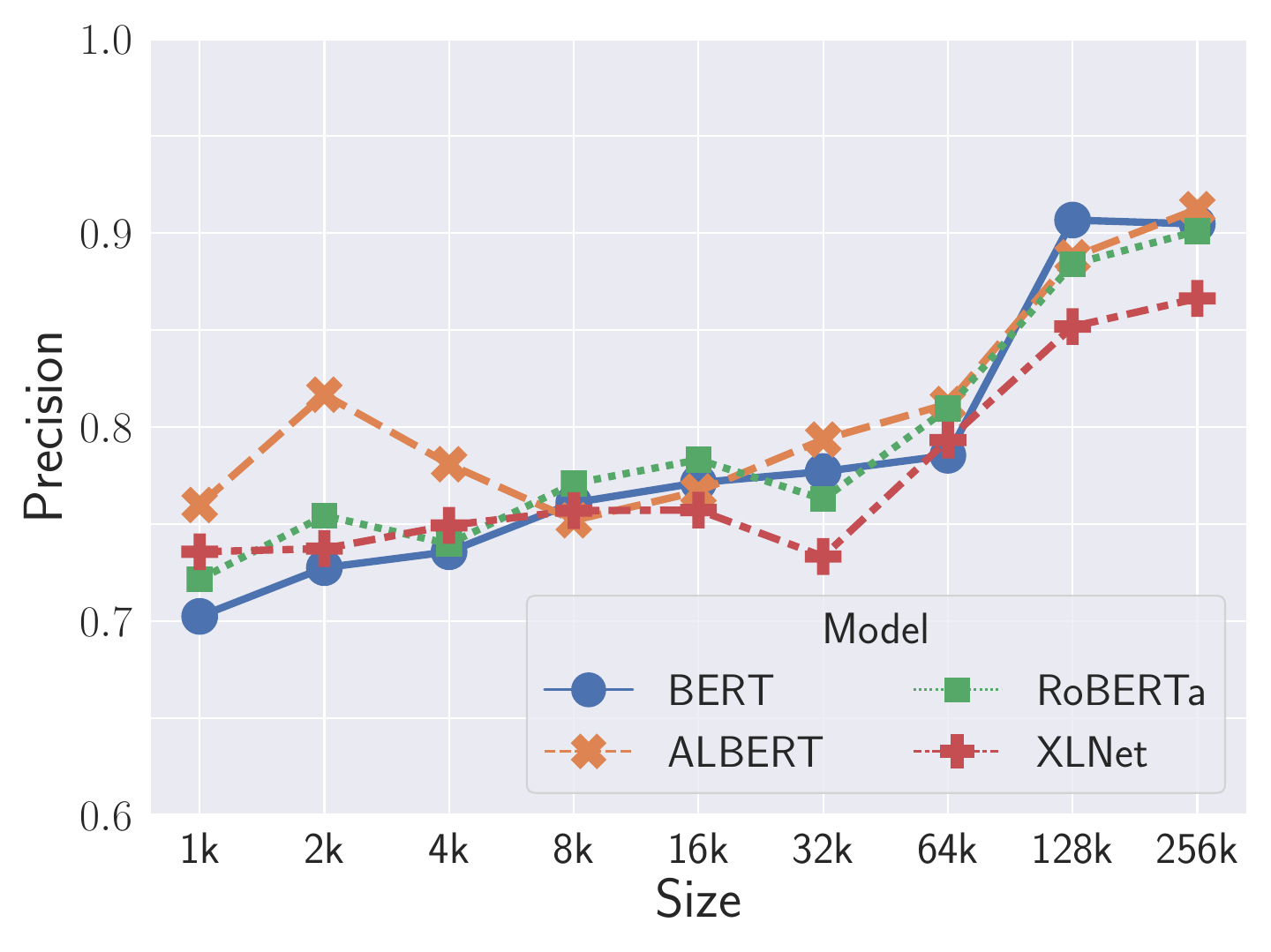}
}\hfill
\subfigure[Recall]{
    \includegraphics[width=0.32\textwidth]{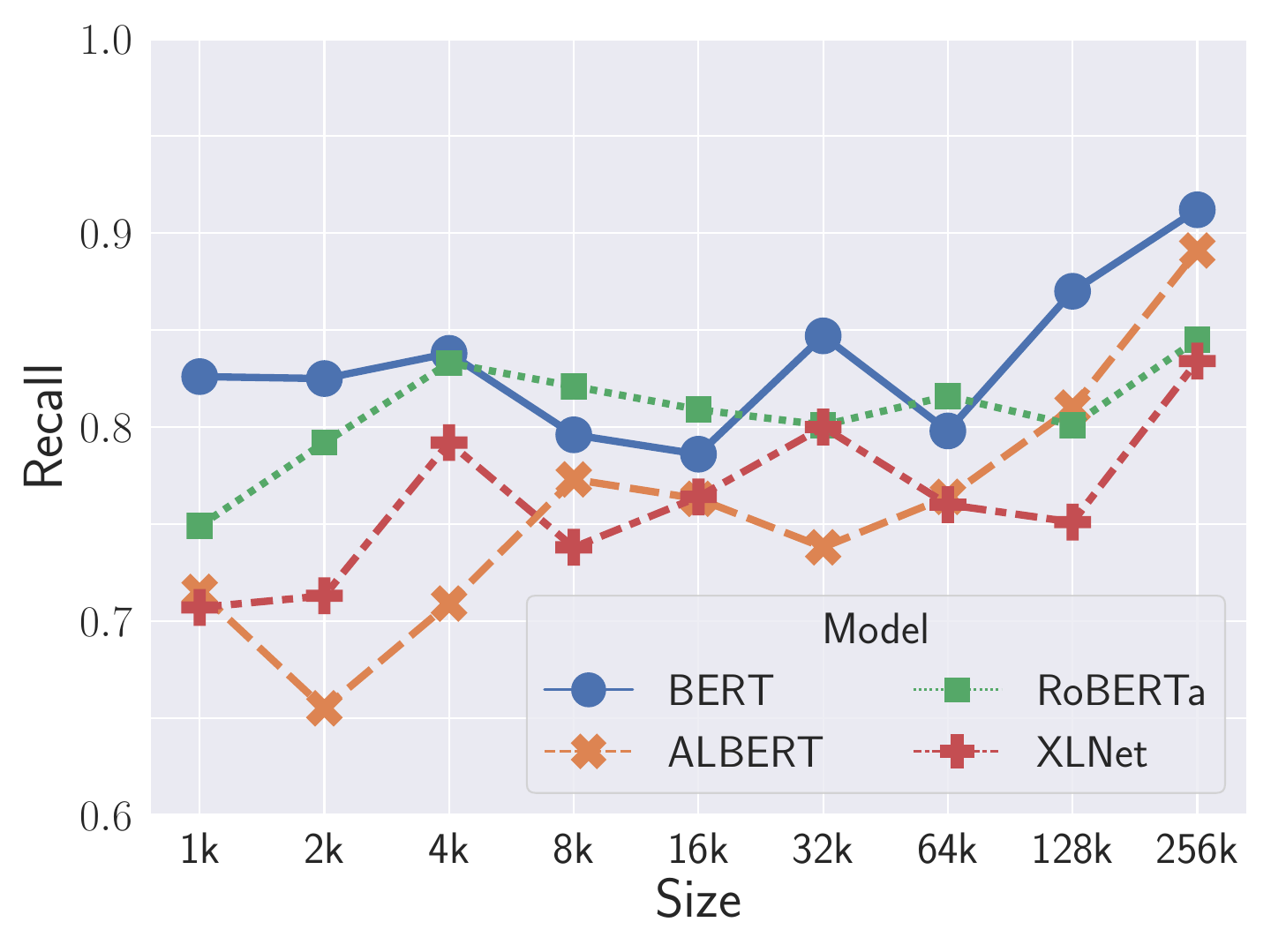}
}
\caption{Attack performance when varying the \textit{size of the $\Spre$} used for training the attack model, with AG's News being the fine-tuning dataset.}
\label{fig:member_size}
\end{figure*}
\begin{figure*}[!t]
\centering 
\subfigure[Accuracy]{
\begin{minipage}[b]{0.32\textwidth}
    \includegraphics[width=1.0\textwidth]{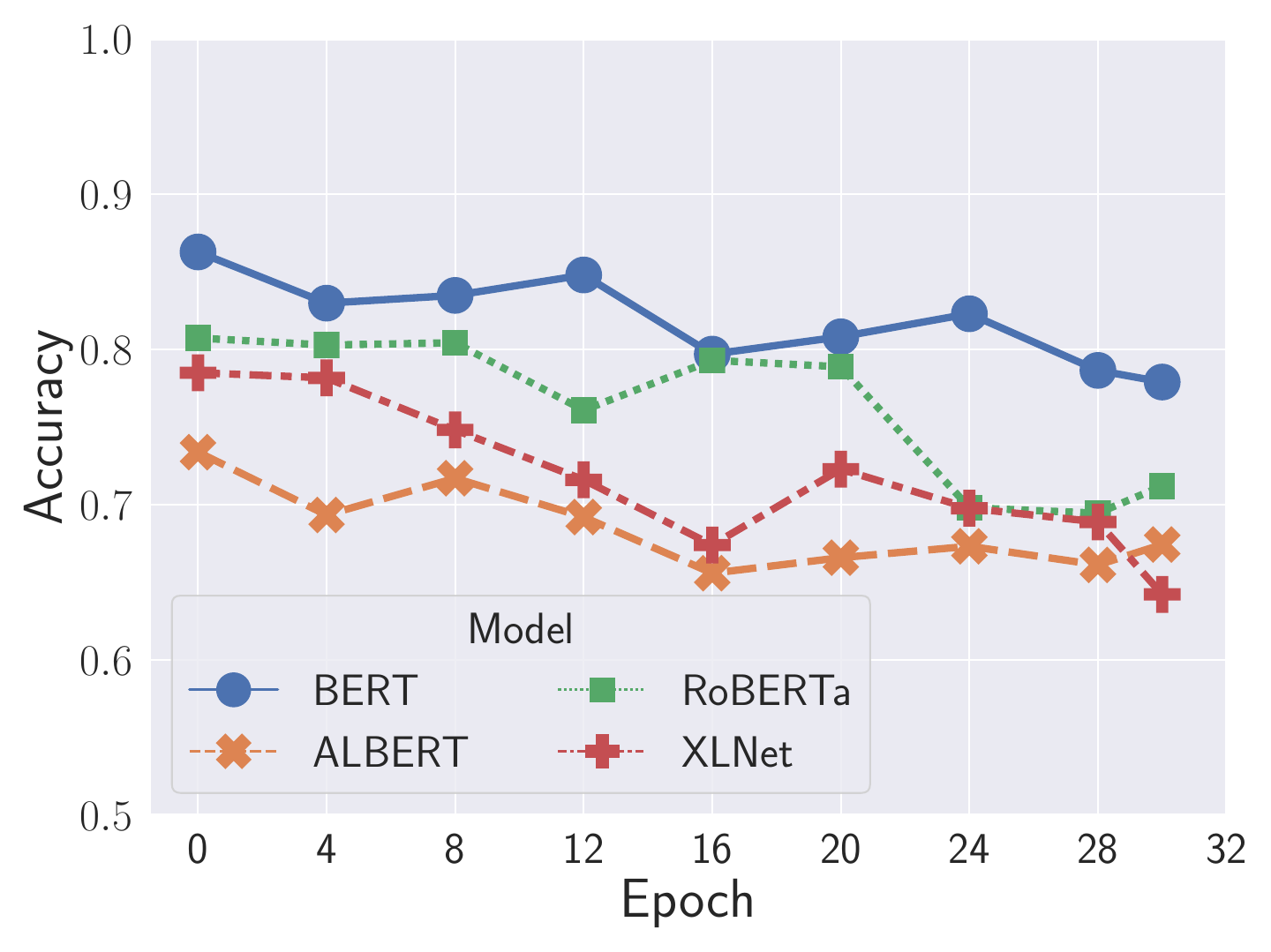}
\end{minipage}
\label{epoch_acc}
}\hfill
\subfigure[Precision]{
\begin{minipage}[b]{0.32\textwidth}
    \includegraphics[width=1.0\textwidth]{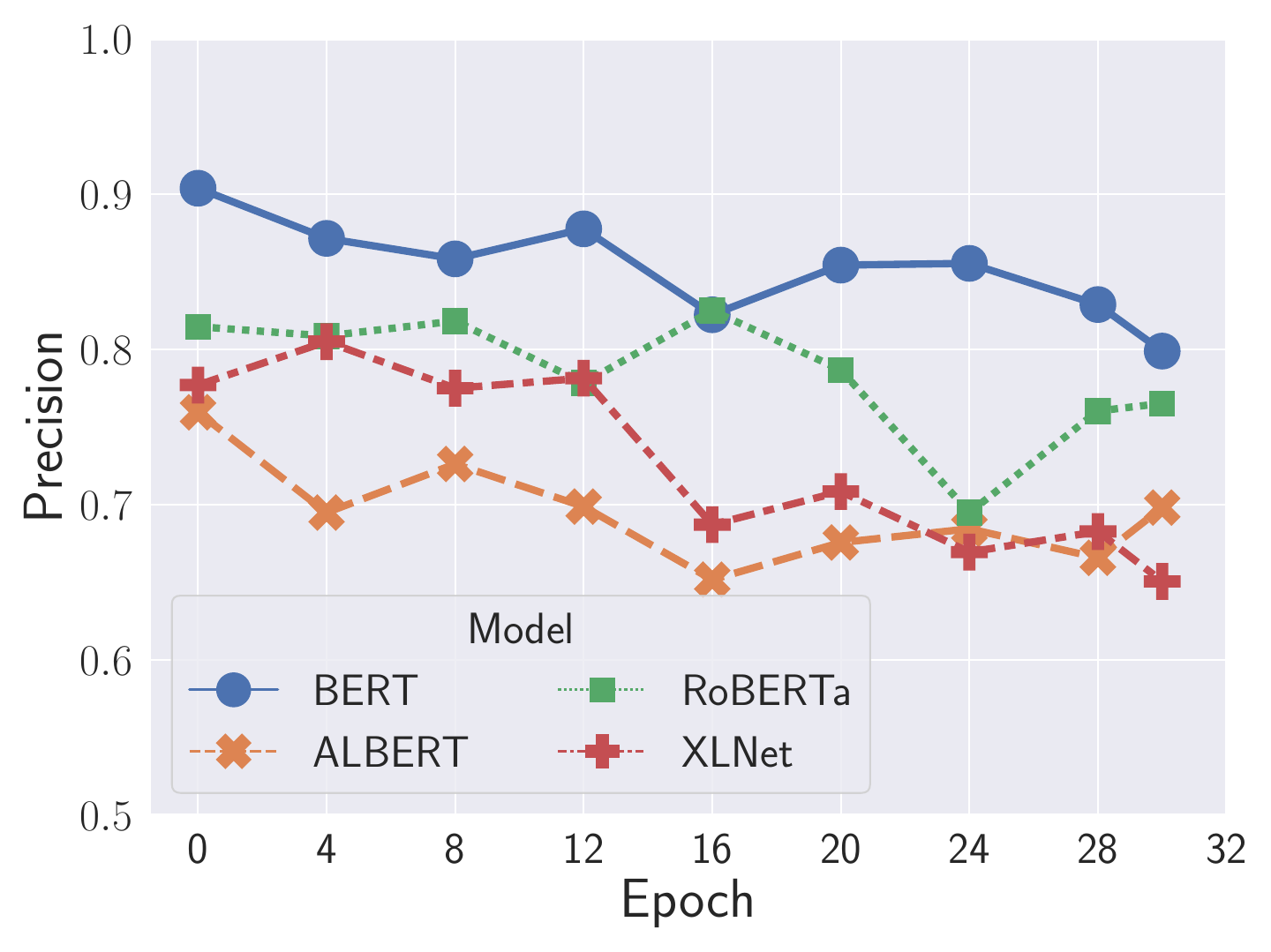}
\end{minipage}
\label{epoch_precision}
}\hfill
\subfigure[Recall]{
\begin{minipage}[b]{0.32\textwidth}
    \includegraphics[width=1.0\textwidth]{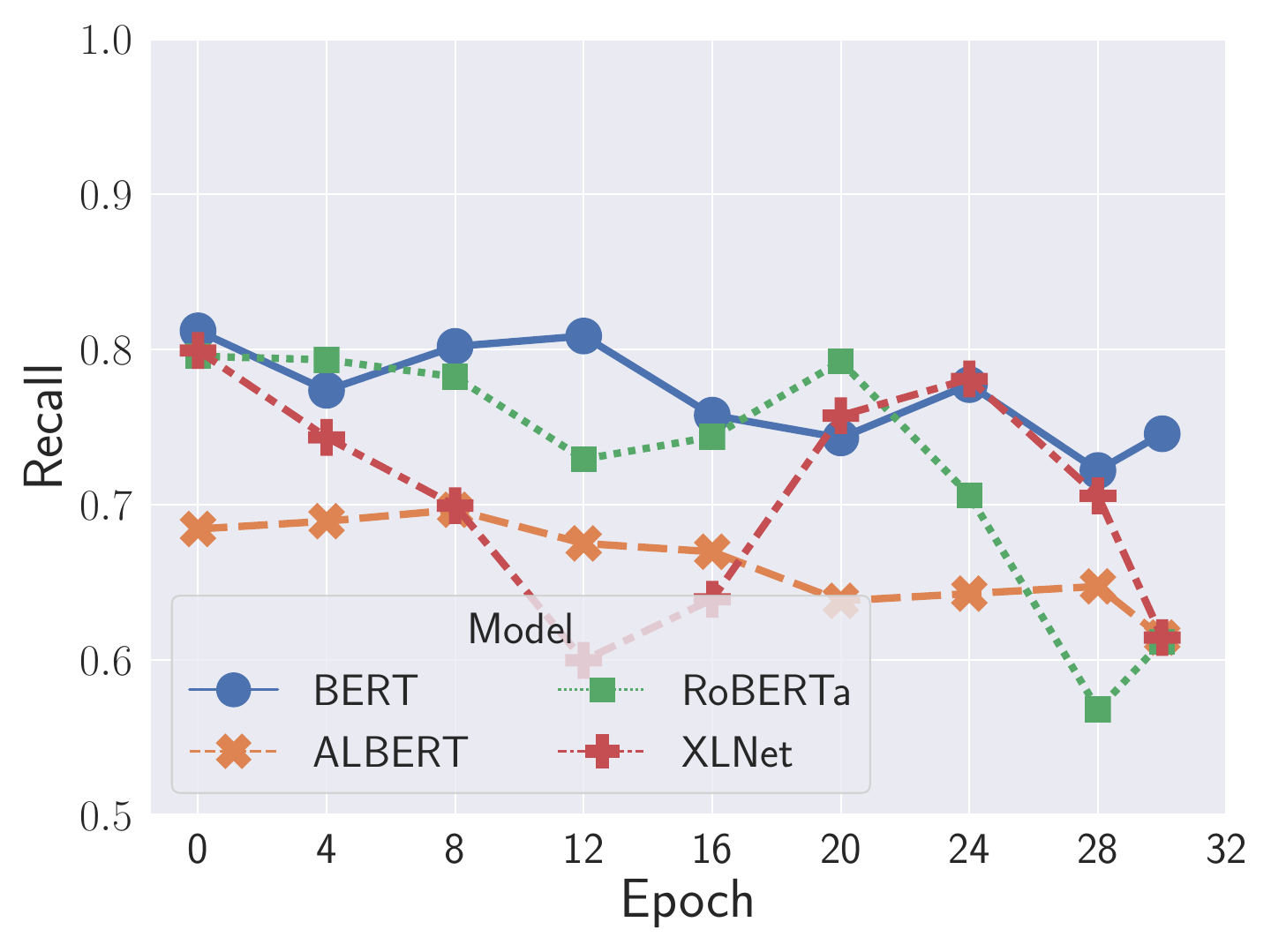}
\end{minipage}
\label{epoch_recall}
}
\caption{Attack performance when varying the \textit{fine-tuning epochs}, with AG's News being the fine-tuninig dataset.}
\label{fig:epoch}
\end{figure*}
\begin{figure*}[!t]
\centering 
\subfigure[Accuracy]{
\begin{minipage}[b]{0.3\textwidth}
    \includegraphics[width=1.0\textwidth]{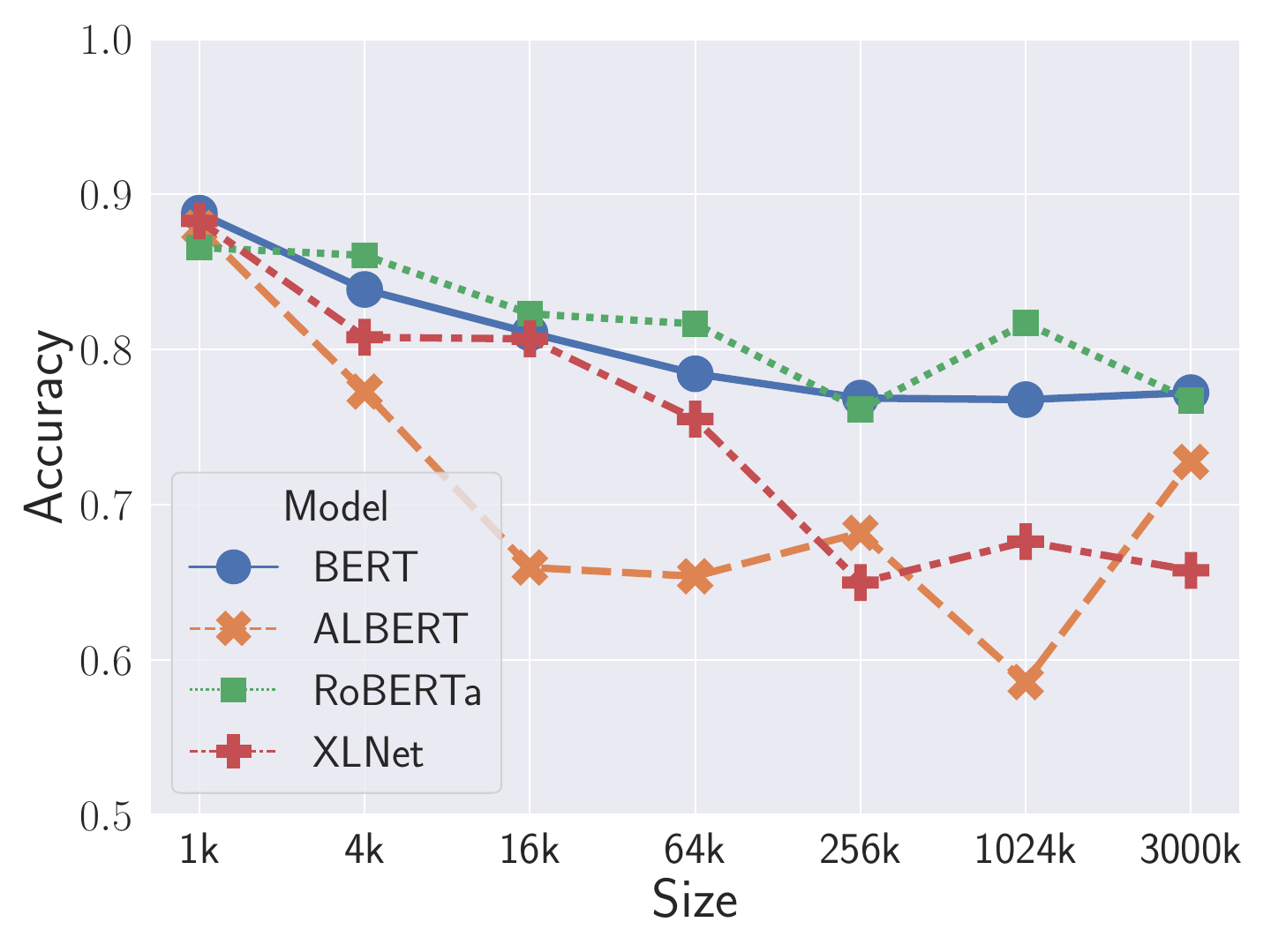}
\end{minipage}

}\hfill
\subfigure[Precision]{
\begin{minipage}[b]{0.3\textwidth}
    \includegraphics[width=1.0\textwidth]{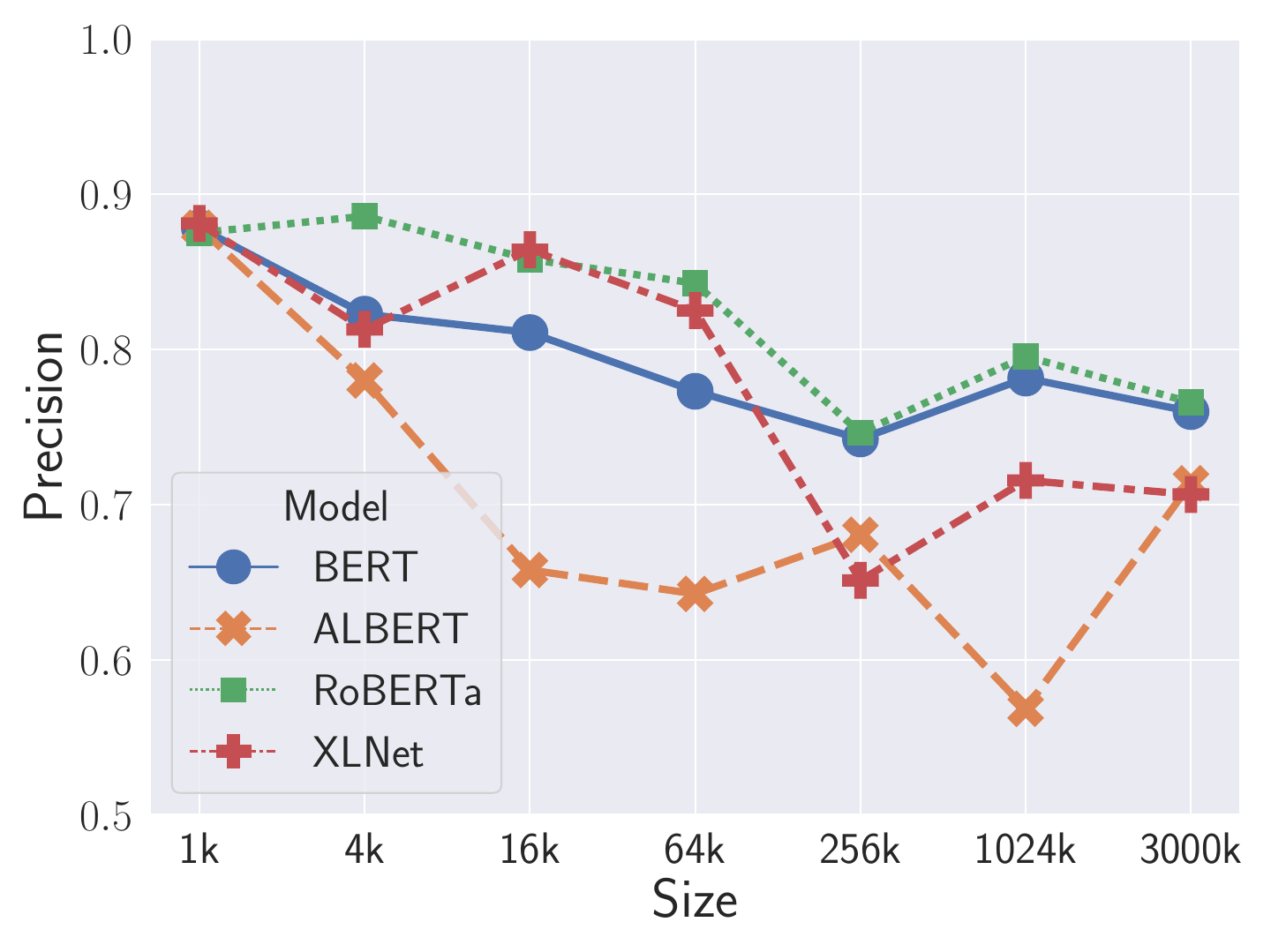}
\end{minipage}

}\hfill
\subfigure[Recall]{
\begin{minipage}[b]{0.3\textwidth}
    \includegraphics[width=1.0\textwidth]{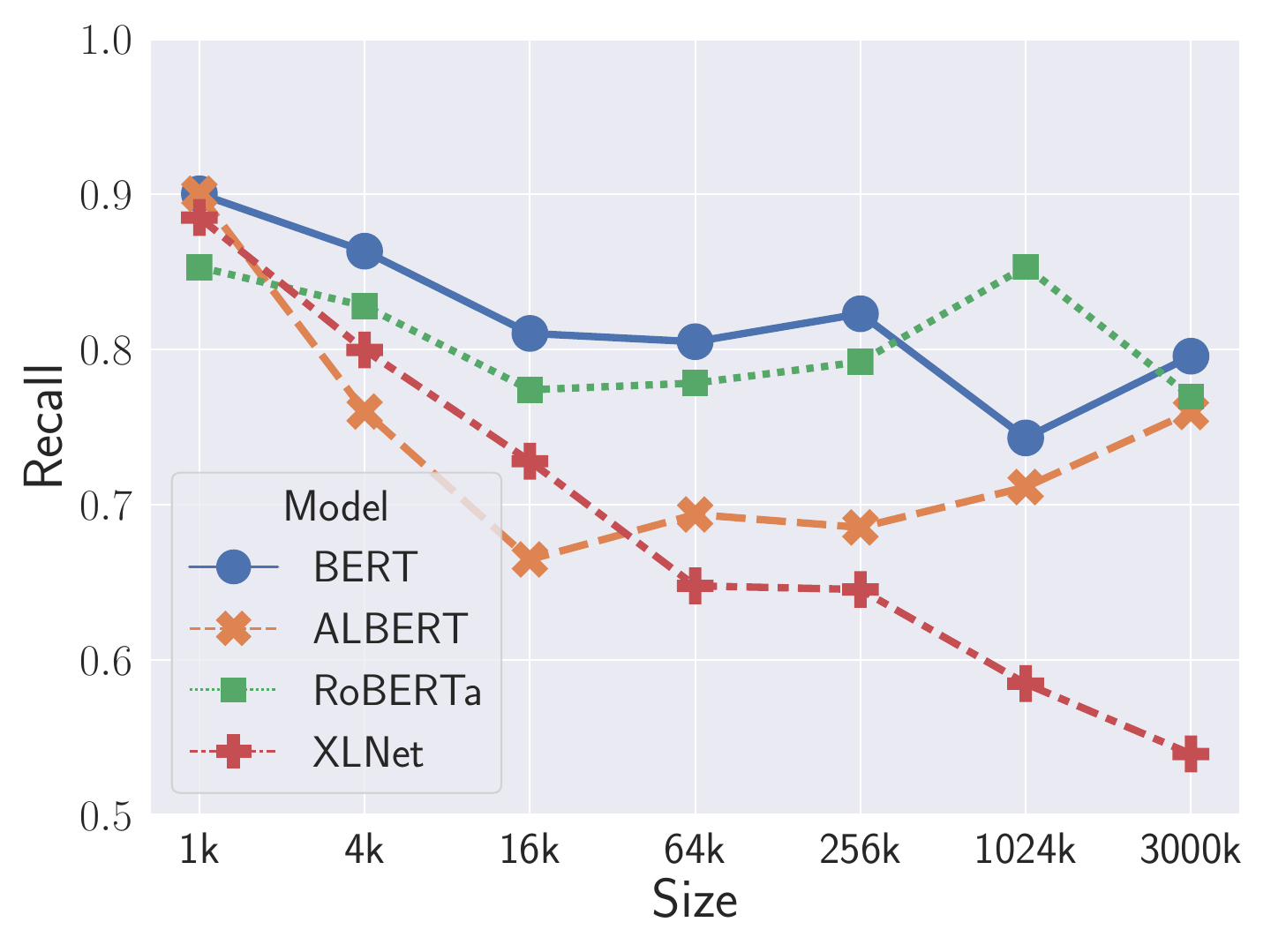}
\end{minipage}
}
\caption{Attack performance when varying the \textit{fine-tuning dataset size}, with Amazon being the fine-tuining dataset.%
}
\label{fig:FT_size}
\vspace{10pt}
\end{figure*}
\subsubsection{Effect of Attack Settings}
\label{sec:experiment:attack_setting}

\paragraph{Attack Training Dataset Relaxation.}
While our standard approach involves utilizing a mixture of available data for robust performance evaluation (as described in \sectionautorefname~\ref{experiment}), we also explore a more demanding scenario for the adversary, where we relax the assumption about the adversary's accessibility to $\Dpre$ and $\Dnon$. Specifically, we consider the adversary only has access to a limited number of samples from a single \textit{pre-training} (termed \textbf{Relaxation-\MakeUppercase{\romannumeral1}}) or \textit{non-member} (termed \textbf{Relaxation-\MakeUppercase{\romannumeral2}}) dataset for training the attack model. Subsequently, the attack performance is evaluated using other unseen pre-training/non-member datasets.
Such relaxations suggest that the adversary possesses only partial knowledge, potentially failing to accurately reflect the complete data distribution. This incomplete representation could introduce disparities between the training and testing data distributions for the attack, thereby adding extra difficulties to its generalization.

We report the attack performance for Relaxation-\MakeUppercase{\romannumeral1} and Relaxation-\MakeUppercase{\romannumeral2} in \figureautorefname~\ref{fig:heatmap_member}. 
Firstly, it is noteworthy that the attack performance remains consistently high, with both precision and recall exceeding 0.7. This high level of performance provides a clear indication of the inherent differences in the target model's responses to the pre-training versus unseen samples, while such differential behavior can generally be delineated by a decision boundary derived from partial knowledge of the distribution.

Moreover, while there exists a noticeable decrease in attack recall when the pre-training datasets  differ for training and testing the attack model, our results still indicate an effective attack, as a MIA is deemed useful as long as the attack can accurately infer a portion of the members~\citep{CCNSTT21} (i.e., maintaining high precision). Specifically, it may be impractical to expect the attack to infer all the members (i.e., aiming for high recall). This is particularly the case when the attack lacks, or has skewed, knowledge about the distribution of the training data.

Additionally, the results underscore the major role of the knowledge about the pre-training data distributions on the attack's success. As these distributions serve as the primary targets of the attacker, their knowledge is crucial to the inference process. Conversely, non-member datasets play a supporting role, aiding the decision-making process but not being the primary focus.

\begin{table*}[!t]
    \centering
    \caption{Attack performance evaluated on \textit{GPT-3.5 generated non-member data}. We report accuracy (A), precision (P), recall (R), and F1-score (F) for each PLE model under different datasets and tasks.}
     \label{table:gpt3.5}
     \vspace{10pt}
    \resizebox{\textwidth}{!}{%
    \begin{tabular}{c|cccc|cccc|cccc|cccc|cccc}
        \toprule
        \multirow{2}{*}{\textbf{Models}} & \multicolumn{4}{c|}{\textbf{SST-2}} & \multicolumn{4}{c|}{\textbf{AG's News}} & \multicolumn{4}{c|}{\textbf{Yelp Review Full}} & \multicolumn{4}{c|}{\textbf{CoNLL2003}} & \multicolumn{4}{c}{\textbf{SQuADv1.0}} \\
        \cline{2-21}
        & A & P & R & F & A & P & R & F & A & P & R & F & A & P & R & F & A & P & R & F \\
        \hline
        BERT &0.56 &0.57 & 0.51& 0.53&  0.61& 0.63& 0.54& 0.58& 0.60& 0.59 & 0.65& 0.62& 0.83& 0.83&0.82 &0.83 &0.83 & 0.91& 0.74&0.81 \\ 
        ALBERT &0.60 &0.63&0.80 &0.70 &0.58 &0.59 & 0.55& 0.57&0.60 & 0.59&0.64 &0.61 & 0.75& 0.74& 0.80&0.76 &0.94 &0.94 & 0.95&0.95 \\
        RoBERTa & 0.57&0.56 &0.64 &0.60 &0.71 & 0.72&0.70 &0.71 & 0.59& 0.58& 0.63& 0.61&0.88 &0.86 &0.92 &0.88 &0.85 &0.83 &0.88 &0.86 \\
        XLNet &0.55 &0.54 &0.79 &0.64 & 0.82& 0.81& 0.84& 0.82&0.80 &0.80 &0.80 & 0.80&0.73 &0.69 & 0.83 & 0.75&0.93 &0.94 & 0.92& 0.93\\
        \bottomrule
    \end{tabular}%
}
\end{table*}

\begin{table*}[!t]
\centering
\caption{
Examples of pre-training data samples and their corresponding GPT-3.5 generated non-member data. The prompt is set to be \textit{"paraphrase the sentence with the same style"}. The examples are categorized based on the attacker's prediction: ``\textbf{\textcolor{mygreen}{correct}}'' when the attacker correctly identifies pre-training data, ``\textbf{\textcolor{myred}{incorrect}}'' when the attacker erroneously identifies the pre-training data sample as unseen. %
}
\label{table:gpt3.5_qualitative}
\vspace{8pt}
\begin{tabularx}{\textwidth}{c|X|X}
\toprule
Type & Pre-training data & Non-member data  \\
\midrule
\multirow{16}{*}{\textbf{\textcolor{mygreen}{correct}}} & in general  the more massive a star is the shorter its lifespan on the main sequence after the hydrogen fuel at the core has been consumed  the star evolves away from the main sequence on the hr diagram  the behavior of a star now depends on its mass  with stars below 0  23 m  becoming white dwarfs. & As a rule, heavier stars spend less time in the main sequence phase. Once they burn through the helium at their core, they move off the main sequence track on the HR diagram. The future of these stars is largely defined by their size, with those below 0.23 solar masses often morphing into red giants.\\
\cline{2-3}
& even though they did n't speak, megan did n't feel awkward around him after what they had done. & not speak anything, Megan was uncomfortable around him after what they had done.\\
\cline{2-3}
& the Oregon Shakespeare Festival, which originally began as a summer outdoor series in Ashland during the 1930s and later moved to Talent, Oregon, to specialize in battery electric motorcycles, has grown to span a season from February to October and incorporate both Shakespearean and non-Shakespearean works. &  Originally launched as a summer indoor event in Medford, Oregon during the 1930s, the Oregon Shakespeare Festival later moved to Talent, Oregon, where it shifted its focus to battery electric motorcycles. It now offers a broad array of performances from March to November, featuring both Shakespearean and contemporary works. \\
 \cline{2-3}
 & april 1774  louis xv fell ill after contracting smallpox and died the following 10 may  the dauphin  louis auguste  succeeded his grandfather as king louis xvi  as eldest brother of the king  louis stanislas received the title monsieur  louis stanislas longed for political influence  he attempted to gain admittance to the king s council in 1774 speculation soared  & In April 1774, Louis XV contracted smallpox and succumbed to the illness on May 10th. His grandson, Louis Auguste, ascended to the throne as King Louis XVI. As the king's eldest brother, Louis Stanislas was bestowed the title "Monsieur." Eager for political clout, he tried to secure a spot in the king's council that same year, amid rising speculation. \\
\hline
\multirow{8}{*}{\textbf{\textcolor{myred}{incorrect}}}  & university in ithaca  new york  he wanted to study the humanities or become an architect like his father  but his father and brother  a scientist  urged him to study a  useful  discipline  as a result  vonnegut majored in biochemistry  but he had little proficiency in the area and was indifferent towards his studies & At a university in Ithaca, New York, he was inclined to pursue humanities or follow in his father's footsteps as an architect. However, under pressure from his father and his brother, who was a scientist, he was persuaded to study a "practical" field. Consequently, Vonnegut ended up majoring in biochemistry \\
\cline{2-3}
& pesh sighed and released her hand. & Pesh let out a sigh and removed her hand.\\
\cline{2-3}
& resignedly, he followed her into the room . &Reluctantly, he trailed her into the room.\\
\cline{2-3}
 & at least not with a man. & Not with a man, at the very least.\\
\bottomrule
\end{tabularx}
\vspace{10pt}
\end{table*}

\paragraph{Effect of Attack Training Set Size.} 
We further investigate the effects of the size of $\Spre$ collected by the adversary for its training. As illustrated in Figure~\ref{fig:member_size}, there exists a clear trend of increasing attack effectiveness as the attack training set size increases. These results align with the expectation that a larger quantity of training data contributes to higher performance in an ML model. Interestingly, we find that even when using only 1k pre-training samples for $\Spre$, the attack performance remains effective, yielding more than 0.7 in terms of precision, recall, and accuracy. Notably, such a training set represents only 0.0077\% of Wikipedia and 0.00067\% of BooksCorpus, indicating a cautionary signal that MIA against PLEs may be much easier to carry out than previously thought.

\subsubsection{Effect of Fine-tuning Settings}
\label{sec:experiment: fine-tuning}

\paragraph{Effect of Fine-tuning Epochs.} 
We report the performance of fine-tuning different epochs in Figure~\ref{fig:epoch}. Generally, as the number of fine-tuning epochs increases, a slight decrease in attack performance is observed. This is primarily due to the more significant alterations in the model parameters of PLEs during the fine-tuning phase under these conditions. As a result, the information about the pre-training data tends to become progressively obfuscated. However, it may still remain vulnerable to potential attacks, as the generic representations derived from the pre-training data often retain their universal utility and are largely preserved.

\paragraph{Effect of Fine-tuning Strategies.}
We present the performance of fine-tuning different parts of a BERT model in Table~\ref{table:partial ft}, reflecting various common fine-tuning strategies adopted in practice. The last column (\textit{utility}) represents the downstream model's classification accuracy on the Yelp dataset. As can be observed, our default setting (i.e., fine-tuning all layers) results in the best downstream utility, which confirms our default configuration is appropriate. Moreover, the other configurations yield similar or even better attack performance. Notably, only updating the word embedding yields the highest attack performance. This aligns with our intuition: such an operation largely preserves the learned semantic information in the remaining layers, making it easier for the attack to extract information about the pre-training dataset.

\begin{table}[!t]
\definecolor{Gray}{gray}{0.9}
\newcommand{\cc}{\cellcolor{Gray}}
\centering
\caption{Comparison of fine-tuning strategies on BERT for the Yelp Review Full downstream task. %
}
\label{table:partial ft}
\vspace{10pt}
\resizebox{\columnwidth}{!}{
\begin{tabular}{l|ccccc}
\toprule
\textbf{Updated Layers} & \textbf{Accuracy} & \textbf{Precision} & \textbf{Recall} & \textbf{F1-score} & \textbf{Utility} \\ 
\midrule
Embedding        & 0.86 & 0.84 & 0.89 & 0.86           & 0.56  \\         
Embedding+Classifier    & 0.84 & 0.84             & 0.83          & 0.85           & 0.61 \\          
Classifier              & 0.83            & 0.82             & 0.84          & 0.83           & 0.47  \\        
\cc All Layers (Default)    & \cc 0.82            & \cc 0.85             & \cc 0.79          & \cc 0.82          & \cc 0.65 \\        \bottomrule 
\end{tabular}}
\end{table}

\paragraph{Effect of the Fine-tuning Dataset Size.}
We illustrate the impact of fine-tuning dataset size in Figure~\ref{fig:FT_size}, where we vary the size of the fine-tuning training set by randomly sampling from the Amazon Review dataset and adopt this data to fine-tune all PLEs examined in this study.
As depicted in Figure~\ref{fig:FT_size}, it appears that attack performance generally diminishes as the size of the fine-tuning data increases. This can be attributed to the fact that introducing more fine-tuning data is akin to augmenting the PLEs with additional knowledge. Consequently, the PLEs will undergo more substantial changes, potentially obscuring their knowledge about the pre-training data during the fine-tuning process, thereby leading to a decrease in attack performance.

\subsubsection{Distribution Similarity Comparison}
\label{sec:experiment:distribution}
To alleviate the potential discrepancy between the pre-training data and unseen data distribution during evaluation, we further conduct experiments by:
\begin{itemize}[topsep=2pt, partopsep=2pt,itemsep=2pt]
    \item selecting an existing dataset as the unseen data (i.e., STORIES) that shares similar semantic styles with the pre-training dataset (i.e., BookCorpus);
    \item adopting a third-party GPT-3.5 to rephrase each pre-training data into a corresponding non-member data sentence, generating a total of 10k non-member sentences while maintaining the same style and semantics.
\end{itemize}

\paragraph{STORIES as the Non-member Data.}
To address concerns that the attack might only distinguish between data distributions, specifically the variations in language style and semantics across datasets, we selected an additional dataset, i.e., STORIES, as the non-member dataset for evaluation. The style of STORIES~\cite{trinh2018simple} closely resembles that of the BookCorpus pre-training datasets: the STORIES dataset was developed by extracting story-like sections from a subset of the CommonCrawl dataset, while BookCorpus consists mostly of fiction books from unpublished authors. Therefore, the two datasets share similar narrative styles and story-like content.As observed in Table~\ref{tab:stories}, the attack performance is consistent with that shown in Figure~\ref{fig:attack_performance}. This further verifies that the success of our data leakage attack against PLEs is not solely attributable to distribution differences between pre-training data and unseen data.

\begin{table}[!h]
\centering
\caption{Attack performance evaluated on STORIES non-member dataset.}
\label{tab:stories}
\vspace{8pt}
\begin{tabular}{@{}l|cccc@{}}
\toprule
\textbf{Downstream Task} & \textbf{Accuracy} & \textbf{Precision} & \textbf{Recall} & \textbf{F1-score} \\ 
\midrule
SST-2 & 0.64 & 0.63 & 0.69 & 0.66\\
AG News & 0.92     & 0.95     & 0.90     & 0.92  \\
Yelp Review Full  & 0.80  & 0.85  & 0.74  & 0.79 \\
NER    & 0.84      & 0.81     & 0.88     & 0.84   \\
QA    & 0.92  & 0.95  & 0.89  & 0.92  \\ 
\bottomrule
\end{tabular}
\end{table}

\paragraph{GPT-3.5 Generated Non-member Data.}
\tableautorefname~\ref{table:gpt3.5} shows that the attack remains generally effective, despite a slight drop in performance compared to the default setting that does not control the semantics of non-member data. 
While comparing different datasets and tasks, we observe that tasks resulting in higher dimensionality of downstream models' outputs tend to make attacks more effective. This aligns with the intuition that higher-dimensional outputs cause downstream models to leak more information about their training data, thereby increasing the threat of attacks (see Appendix for additional experiments).
Additionally, the qualitative examples presented in \tableautorefname~\ref{table:gpt3.5_qualitative} reveal some potentially interesting trends. In general, the attack demonstrates higher effectiveness with more confident predictions for longer sentences containing meaningful entities that provide more fine-grained information and are indeed more privacy-sensitive. Conversely, samples for which the attack is uncertain of their membership status tend to feature 'neutral' text that could generally occur in everyday life. These observations indicate the potential threat and highlight the need for careful sanitization of (the informative or sensitive entities contained within) pre-training data when deploying PLEs to ensure compliance with privacy regulations.

\noindent\paragraph{Quantitative Measure of Distribution Similarity.}
We further quantify the distribution similarity using common standard metrics including: BERTScore~\footnote{\url{https://huggingface.co/spaces/evaluate-metric/bertscore}}, RougeScore~\footnote{\url{https://huggingface.co/spaces/evaluate-metric/rouge}}, Fréchet inception distance~(FID)~\footnote{\url{https://pytorch.org/ignite/generated/ignite.metrics.FID.html}} and Maximum Mean Discrepancy (MMD)~\footnote{\url{https://lightning.ai/docs/torchmetrics/stable/image/kernel_inception_distance.html}}. For BERTScore and RougeScore, higher values signify greater similarity. Conversely, for FID and MMD, lower values indicate higher similarity. We quantified the distribution similarity between the pre-training (Wikipedia) and two different types of non-member data:
\begin{itemize}[topsep=2pt,itemsep=2pt]
    \item \textbf{Random Subset}: random selections from the same pre-training dataset (i.e., Wikipedia).
    \item 
\textbf{GPT Non-member}: data generated by rephrasing the pre-training data using a GPT-3.5 model.
\end{itemize}
 Both types are compared against the same reference pre-training set for fair comparison. Notably, \tableautorefname~\ref{distribution similarity} shows that the GPT-generated non-member data achieve a remarkable level of distribution similarity with pre-training data, even surpassing that of random subsets within the same pre-training dataset. 
Such findings underscore that our successful detection of pre-training data leakage is most likely not merely due to divergent distributions between pre-training and non-member data. 

\begin{table}[!h]
    \centering
        \caption{Quantification of distribution similarity. }
    \label{distribution similarity}
    \vspace{8pt}
     \resizebox{0.45\textwidth}{!}{
    \begin{tabular}{l|cc}
        \toprule
        \textbf{Similarity Metrics}& Random Subset & GPT-3.5 Generated  \\
        \midrule
        BERTScore ($\uparrow$) & 0.80 & \textbf{0.91} \\
        RougeScore ($\uparrow$) & 0.15 & \textbf{0.48}  \\
        FID ($\downarrow$) & 9.39 & \textbf{3.45}  \\
        MMD ($\downarrow$) & 192 & \textbf{67}  \\
        \bottomrule
    \end{tabular}
    }

\end{table}

\section{Conclusion}
In this work, we pioneer the systematic study of potential data leakage associated with PLEs. Specifically, we consider a realistic and challenging scenario where the adversary can only gain access to the output of downstream models adapted from PLEs. We conduct extensive and rigorous evaluations that span a variety of PLE architectures, different types of downstream tasks, and a number of important factors that affect membership leakage from different angles. Our experimental results yield intriguing findings, suggesting that what appears to be a safe usage scenario might indeed be problematic, presenting a privacy threat to PLEs that is far greater than previously believed. Lastly, our in-depth analysis, together with key insights, raises critical considerations for future model developers. %

\clearpage
\newpage
\section*{Ethics Statement}
Our work contributes to a better understanding of the potential threats associated with the usage of pre-trained language encoders, providing insights that raise awareness and anticipate positive societal impacts. %
We are not aware of any additional negative societal impacts beyond the generic risks of ML technology.
\vspace{-10pt}

\bibliography{iclr23}

\begin{thebibliography}{42}
\providecommand{\natexlab}[1]{#1}
\providecommand{\url}[1]{\texttt{#1}}
\expandafter\ifx\csname urlstyle\endcsname\relax
  \providecommand{\doi}[1]{doi: #1}\else
  \providecommand{\doi}{doi: \begingroup \urlstyle{rm}\Url}\fi

\bibitem[Bordes et~al.(2014)Bordes, Chopra, and Weston]{BCW14}
A.~Bordes, S.~Chopra, and J.~Weston.
\newblock {Question Answering with Subgraph Embeddings}.
\newblock In \emph{{Conference on Empirical Methods in Natural Language
  Processing (EMNLP)}}, pages 615--620. ACL, 2014.

\bibitem[Brown et~al.(2020{\natexlab{a}})Brown, Mann, Ryder, Subbiah, Kaplan,
  Dhariwal, Neelakantan, Shyam, Sastry, Askell, et~al.]{brown2020language}
T.~Brown, B.~Mann, N.~Ryder, M.~Subbiah, J.~D. Kaplan, P.~Dhariwal,
  A.~Neelakantan, P.~Shyam, G.~Sastry, A.~Askell, et~al.
\newblock Language models are few-shot learners.
\newblock \emph{nips}, 33:\penalty0 1877--1901, 2020{\natexlab{a}}.

\bibitem[Brown et~al.(2020{\natexlab{b}})Brown, Mann, Ryder, Subbiah, Kaplan,
  Dhariwal, Neelakantan, Shyam, Sastry, Askell, Agarwal, Herbert{-}Voss,
  Krueger, Henighan, Child, Ramesh, Ziegler, Wu, Winter, Hesse, Chen, Sigler,
  Litwin, Gray, Chess, Clark, Berner, McCandlish, Radford, Sutskever, and
  Amodei]{BMRSKDNSSAAHKHCRZWWHCSLGCCBMRSA20}
T.~B. Brown, B.~Mann, N.~Ryder, M.~Subbiah, J.~Kaplan, P.~Dhariwal,
  A.~Neelakantan, P.~Shyam, G.~Sastry, A.~Askell, S.~Agarwal,
  A.~Herbert{-}Voss, G.~Krueger, T.~Henighan, R.~Child, A.~Ramesh, D.~M.
  Ziegler, J.~Wu, C.~Winter, C.~Hesse, M.~Chen, E.~Sigler, M.~Litwin, S.~Gray,
  B.~Chess, J.~Clark, C.~Berner, S.~McCandlish, A.~Radford, I.~Sutskever, and
  D.~Amodei.
\newblock {Language Models are Few-Shot Learners}.
\newblock In \emph{{Annual Conference on Neural Information Processing Systems
  (NeurIPS)}}. NeurIPS, 2020{\natexlab{b}}.

\bibitem[Carlini et~al.(2021{\natexlab{a}})Carlini, Chien, Nasr, Song, Terzis,
  and Tram{\`{e}}r]{CCNSTT21}
N.~Carlini, S.~Chien, M.~Nasr, S.~Song, A.~Terzis, and F.~Tram{\`{e}}r.
\newblock {Membership Inference Attacks From First Principles}.
\newblock \emph{{CoRR abs/2112.03570}}, 2021{\natexlab{a}}.

\bibitem[Carlini et~al.(2021{\natexlab{b}})Carlini, Tramer, Wallace, Jagielski,
  Herbert-Voss, Lee, Roberts, Brown, Song, Erlingsson,
  et~al.]{carlini2021extracting}
N.~Carlini, F.~Tramer, E.~Wallace, M.~Jagielski, A.~Herbert-Voss, K.~Lee,
  A.~Roberts, T.~B. Brown, D.~Song, U.~Erlingsson, et~al.
\newblock Extracting training data from large language models.
\newblock In \emph{USENIX Security Symposium}, volume~6, 2021{\natexlab{b}}.

\bibitem[Choo et~al.(2021)Choo, Tram{\`e}r, Carlini, and Papernot]{CTCP21}
C.~A.~C. Choo, F.~Tram{\`e}r, N.~Carlini, and N.~Papernot.
\newblock {Label-Only Membership Inference Attacks}.
\newblock In \emph{{International Conference on Machine Learning (ICML)}},
  pages 1964--1974. PMLR, 2021.

\bibitem[Devlin et~al.(2019)Devlin, Chang, Lee, and Toutanova]{DCLT19}
J.~Devlin, M.~Chang, K.~Lee, and K.~Toutanova.
\newblock {BERT: Pre-training of Deep Bidirectional Transformers for Language
  Understanding}.
\newblock In \emph{{Conference of the North American Chapter of the Association
  for Computational Linguistics: Human Language Technologies (NAACL-HLT)}},
  pages 4171--4186. ACL, 2019.

\bibitem[Goodfellow et~al.(2013)Goodfellow, Mirza, Xiao, Courville, and
  Bengio]{goodfellow2013empirical}
I.~J. Goodfellow, M.~Mirza, D.~Xiao, A.~Courville, and Y.~Bengio.
\newblock An empirical investigation of catastrophic forgetting in
  gradient-based neural networks.
\newblock \emph{arXiv preprint arXiv:1312.6211}, 2013.

\bibitem[He et~al.(2022{\natexlab{a}})He, Li, Xu, Cornelius, and
  Zhang]{HLXCZ22}
X.~He, Z.~Li, W.~Xu, C.~Cornelius, and Y.~Zhang.
\newblock {Membership-Doctor: Comprehensive Assessment of Membership Inference
  Against Machine Learning Models}.
\newblock \emph{{CoRR abs/2208.10445}}, 2022{\natexlab{a}}.

\bibitem[He et~al.(2022{\natexlab{b}})He, Liu, Gong, and Zhang]{HLGZ22}
X.~He, H.~Liu, N.~Z. Gong, and Y.~Zhang.
\newblock {Semi-Leak: Membership Inference Attacks Against Semi-supervised
  Learning}.
\newblock In \emph{{European Conference on Computer Vision (ECCV)}}. Springer,
  2022{\natexlab{b}}.

\bibitem[Hu et~al.(2021)Hu, Salcic, Sun, Dobbie, Yu, and Zhang]{HSSDYZ21}
H.~Hu, Z.~Salcic, L.~Sun, G.~Dobbie, P.~S. Yu, and X.~Zhang.
\newblock {Membership Inference Attacks on Machine Learning: A Survey}.
\newblock \emph{{ACM Computing Surveys}}, 2021.

\bibitem[Jagannatha et~al.(2021)Jagannatha, Rawat, and Yu]{JRY21}
A.~Jagannatha, B.~P.~S. Rawat, and H.~Yu.
\newblock {Membership Inference Attack Susceptibility of Clinical Language
  Models}.
\newblock \emph{{CoRR abs/2104.08305}}, 2021.

\bibitem[Kingma and Ba(2015)]{KB15}
D.~P. Kingma and J.~Ba.
\newblock {Adam: A Method for Stochastic Optimization}.
\newblock In \emph{{International Conference on Learning Representations
  (ICLR)}}, 2015.

\bibitem[Lan et~al.(2020)Lan, Chen, Goodman, Gimpel, Sharma, and
  Soricut]{LCGGSS20}
Z.~Lan, M.~Chen, S.~Goodman, K.~Gimpel, P.~Sharma, and R.~Soricut.
\newblock {{ALBERT:} {A} Lite {BERT} for Self-supervised Learning of Language
  Representations}.
\newblock In \emph{{International Conference on Learning Representations
  (ICLR)}}, 2020.

\bibitem[Lee et~al.(2020)Lee, Yoon, Kim, Kim, Kim, So, and Kang]{LYKKKSK20}
J.~Lee, W.~Yoon, S.~Kim, D.~Kim, S.~Kim, C.~H. So, and J.~Kang.
\newblock {BioBERT: a pre-trained biomedical language representation model for
  biomedical text mining}.
\newblock \emph{{Bioinformatics}}, 2020.

\bibitem[Lewis et~al.(2020)Lewis, Liu, Goyal, Ghazvininejad, Mohamed, Levy,
  Stoyanov, and Zettlemoyer]{LLGGMLSZ20}
M.~Lewis, Y.~Liu, N.~Goyal, M.~Ghazvininejad, A.~Mohamed, O.~Levy, V.~Stoyanov,
  and L.~Zettlemoyer.
\newblock {BART: Denoising Sequence-to-Sequence Pre-training for Natural
  Language Generation, Translation, and Comprehension}.
\newblock In \emph{{Annual Meeting of the Association for Computational
  Linguistics (ACL)}}, pages 7871--7880. ACL, 2020.

\bibitem[Li et~al.(2020)Li, Feng, Meng, Han, Wu, and Li]{LFMHWL20}
X.~Li, J.~Feng, Y.~Meng, Q.~Han, F.~Wu, and J.~Li.
\newblock {A Unified {MRC} Framework for Named Entity Recognition}.
\newblock In \emph{{Annual Meeting of the Association for Computational
  Linguistics (ACL)}}, pages 5849--5859. ACL, 2020.

\bibitem[Li et~al.(2022)Li, Liu, He, Yu, Backes, and Zhang]{LLHYBZ222}
Z.~Li, Y.~Liu, X.~He, N.~Yu, M.~Backes, and Y.~Zhang.
\newblock {Auditing Membership Leakages of Multi-Exit Networks}.
\newblock \emph{{CoRR abs/2208.11180}}, 2022.

\bibitem[Liu et~al.(2019)Liu, Ott, Goyal, Du, Joshi, Chen, Levy, Lewis,
  Zettlemoyer, and Stoyanov]{LOGDJCLLZS19}
Y.~Liu, M.~Ott, N.~Goyal, J.~Du, M.~Joshi, D.~Chen, O.~Levy, M.~Lewis,
  L.~Zettlemoyer, and V.~Stoyanov.
\newblock {RoBERTa: {A} Robustly Optimized {BERT} Pretraining Approach}.
\newblock \emph{{CoRR abs/1907.11692}}, 2019.

\bibitem[Maini et~al.(2021)Maini, Yaghini, and Papernot]{maini2021dataset}
P.~Maini, M.~Yaghini, and N.~Papernot.
\newblock Dataset inference: Ownership resolution in machine learning.
\newblock In \emph{iclr}, 2021.

\bibitem[Mireshghallah et~al.(2022{\natexlab{a}})Mireshghallah, Goyal, Uniyal,
  Berg{-}Kirkpatrick, and Shokri]{MGUBS22}
F.~Mireshghallah, K.~Goyal, A.~Uniyal, T.~Berg{-}Kirkpatrick, and R.~Shokri.
\newblock {Quantifying Privacy Risks of Masked Language Models Using Membership
  Inference Attacks}.
\newblock \emph{{CoRR abs/2203.03929}}, 2022{\natexlab{a}}.

\bibitem[Mireshghallah et~al.(2022{\natexlab{b}})Mireshghallah, Uniyal, Wang,
  Evans, and Berg-Kirkpatrick]{mireshghallah2022memorization}
F.~Mireshghallah, A.~Uniyal, T.~Wang, D.~Evans, and T.~Berg-Kirkpatrick.
\newblock Memorization in nlp fine-tuning methods.
\newblock \emph{arXiv preprint arXiv:2205.12506}, 2022{\natexlab{b}}.

\bibitem[Munikar et~al.(2019)Munikar, Shakya, and Shrestha]{MSS19}
M.~Munikar, S.~Shakya, and A.~Shrestha.
\newblock {Fine-grained Sentiment Classification using BERT}.
\newblock \emph{{CoRR abs/1910.03474}}, 2019.

\bibitem[Pang and Lee(2005)]{pang2005seeing}
B.~Pang and L.~Lee.
\newblock Seeing stars: Exploiting class relationships for sentiment
  categorization with respect to rating scales.
\newblock \emph{arXiv preprint cs/0506075}, 2005.

\bibitem[Rahimian et~al.(2020)Rahimian, Orekondy, and Fritz]{ROF202}
S.~Rahimian, T.~Orekondy, and M.~Fritz.
\newblock {Sampling Attacks: Amplification of Membership Inference Attacks by
  Repeated Queries}.
\newblock \emph{{CoRR abs/2009.00395}}, 2020.

\bibitem[{Rajpurkar} et~al.(2016){Rajpurkar}, {Zhang}, {Lopyrev}, and
  {Liang}]{2016arXiv160605250R}
P.~{Rajpurkar}, J.~{Zhang}, K.~{Lopyrev}, and P.~{Liang}.
\newblock {SQuAD: 100,000+ Questions for Machine Comprehension of Text}.
\newblock \emph{arXiv e-prints}, art. arXiv:1606.05250, 2016.

\bibitem[Ramponi and Plank(2020)]{ramponi2020neural}
A.~Ramponi and B.~Plank.
\newblock Neural unsupervised domain adaptation in nlp---a survey.
\newblock \emph{arXiv preprint arXiv:2006.00632}, 2020.

\bibitem[Salem et~al.(2019)Salem, Zhang, Humbert, Berrang, Fritz, and
  Backes]{SZHBFB19}
A.~Salem, Y.~Zhang, M.~Humbert, P.~Berrang, M.~Fritz, and M.~Backes.
\newblock {ML-Leaks: Model and Data Independent Membership Inference Attacks
  and Defenses on Machine Learning Models}.
\newblock In \emph{ndss}. Internet Society, 2019.

\bibitem[Shejwalkar et~al.(2021)Shejwalkar, Inan, Houmansadr, and Sim]{SIHS21}
V.~Shejwalkar, H.~A. Inan, A.~Houmansadr, and R.~Sim.
\newblock {Membership Inference Attacks Against NLP Classification Models}.
\newblock In \emph{{PriML Workshop (PriML)}}. NeurIPS, 2021.

\bibitem[Shen et~al.(2021)Shen, Ji, Zhang, Li, Chen, Shi, Fang, Yin, and
  Wang]{SJZLCSFYW21}
L.~Shen, S.~Ji, X.~Zhang, J.~Li, J.~Chen, J.~Shi, C.~Fang, J.~Yin, and T.~Wang.
\newblock {Backdoor Pre-trained Models Can Transfer to All}.
\newblock In \emph{{ACM SIGSAC Conference on Computer and Communications
  Security (CCS)}}, pages 3141--3158. ACM, 2021.

\bibitem[Shokri et~al.(2017)Shokri, Stronati, Song, and Shmatikov]{SSSS17}
R.~Shokri, M.~Stronati, C.~Song, and V.~Shmatikov.
\newblock {Membership Inference Attacks Against Machine Learning Models}.
\newblock In \emph{{IEEE Symposium on Security and Privacy (S\&P)}}, pages
  3--18. IEEE, 2017.

\bibitem[Socher et~al.(2013)Socher, Perelygin, Wu, Chuang, Manning, Ng, and
  Potts]{SPWCMNP13}
R.~Socher, A.~Perelygin, J.~Wu, J.~Chuang, C.~D. Manning, A.~Y. Ng, and
  C.~Potts.
\newblock {Recursive Deep Models for Semantic Compositionality Over a Sentiment
  Treebank}.
\newblock In \emph{{Conference on Empirical Methods in Natural Language
  Processing (EMNLP)}}, pages 1631--1642. ACL, 2013.

\bibitem[Song and Shmatikov(2019)]{SS19}
C.~Song and V.~Shmatikov.
\newblock {Auditing Data Provenance in Text-Generation Models}.
\newblock In \emph{{ACM Conference on Knowledge Discovery and Data Mining
  (KDD)}}, pages 196--206. ACM, 2019.

\bibitem[Sun et~al.(2019)Sun, Qiu, Xu, and Huang]{SQXH19}
C.~Sun, X.~Qiu, Y.~Xu, and X.~Huang.
\newblock {How to Fine-Tune BERT for Text Classification?}
\newblock In \emph{{China National Conference on Chinese Computational
  Linguistics (CCL)}}, pages 194--206. Springer, 2019.

\bibitem[Tirumala et~al.(2022)Tirumala, Markosyan, Zettlemoyer, and
  Aghajanyan]{tirumala2022memorization}
K.~Tirumala, A.~Markosyan, L.~Zettlemoyer, and A.~Aghajanyan.
\newblock Memorization without overfitting: Analyzing the training dynamics of
  large language models.
\newblock \emph{nips}, 35:\penalty0 38274--38290, 2022.

\bibitem[Touvron et~al.(2023)Touvron, Lavril, Izacard, Martinet, Lachaux,
  Lacroix, Rozi{\`e}re, Goyal, Hambro, Azhar, et~al.]{touvron2023llama}
H.~Touvron, T.~Lavril, G.~Izacard, X.~Martinet, M.-A. Lachaux, T.~Lacroix,
  B.~Rozi{\`e}re, N.~Goyal, E.~Hambro, F.~Azhar, et~al.
\newblock Llama: Open and efficient foundation language models.
\newblock \emph{arXiv preprint arXiv:2302.13971}, 2023.

\bibitem[Trinh and Le(2018)]{trinh2018simple}
T.~H. Trinh and Q.~V. Le.
\newblock A simple method for commonsense reasoning.
\newblock \emph{arXiv preprint arXiv:1806.02847}, 2018.

\bibitem[Vaswani et~al.(2017)Vaswani, Shazeer, Parmar, Uszkoreit, Jones, Gomez,
  Kaiser, and Polosukhin]{VSPUJGKP17}
A.~Vaswani, N.~Shazeer, N.~Parmar, J.~Uszkoreit, L.~Jones, A.~N. Gomez,
  L.~Kaiser, and I.~Polosukhin.
\newblock {Attention is All you Need}.
\newblock In \emph{{Annual Conference on Neural Information Processing Systems
  (NIPS)}}, pages 5998--6008. NIPS, 2017.

\bibitem[Wolf et~al.(2019)Wolf, Debut, Sanh, Chaumond, Delangue, Moi, Cistac,
  Rault, Louf, Funtowicz, and Brew]{WDSCDMCRLFB19}
T.~Wolf, L.~Debut, V.~Sanh, J.~Chaumond, C.~Delangue, A.~Moi, P.~Cistac,
  T.~Rault, R.~Louf, M.~Funtowicz, and J.~Brew.
\newblock {HuggingFace's Transformers: State-of-the-art Natural Language
  Processing}.
\newblock \emph{{CoRR abs/1910.03771}}, 2019.

\bibitem[Yang et~al.(2019)Yang, Dai, Yang, Carbonell, Salakhutdinov, and
  Le]{YDYCSL19}
Z.~Yang, Z.~Dai, Y.~Yang, J.~G. Carbonell, R.~Salakhutdinov, and Q.~V. Le.
\newblock {XLNet: Generalized Autoregressive Pretraining for Language
  Understanding}.
\newblock In \emph{{Annual Conference on Neural Information Processing Systems
  (NeurIPS)}}. NeurIPS, 2019.

\bibitem[Ye et~al.(2021)Ye, Maddi, Murakonda, and Shokri]{YMMS21}
J.~Ye, A.~Maddi, S.~K. Murakonda, and R.~Shokri.
\newblock {Enhanced Membership Inference Attacks against Machine Learning
  Models}.
\newblock \emph{{CoRR abs/2111.09679}}, 2021.

\bibitem[Zhang et~al.(2015)Zhang, Zhao, and LeCun]{ZZL15}
X.~Zhang, J.~Zhao, and Y.~LeCun.
\newblock {Character-level Convolutional Networks for Text Classification}.
\newblock In \emph{{Annual Conference on Neural Information Processing Systems
  (NIPS)}}, pages 649--657. NIPS, 2015.

\end{thebibliography}
\clearpage
{\noindent\LARGE \textbf{Appendix}}
\appendix

\begin{figure*}[t]
\centering 
\subfigure[Accuracy]{
\begin{minipage}[b]{0.32\textwidth}
    \includegraphics[width=1.0\textwidth]{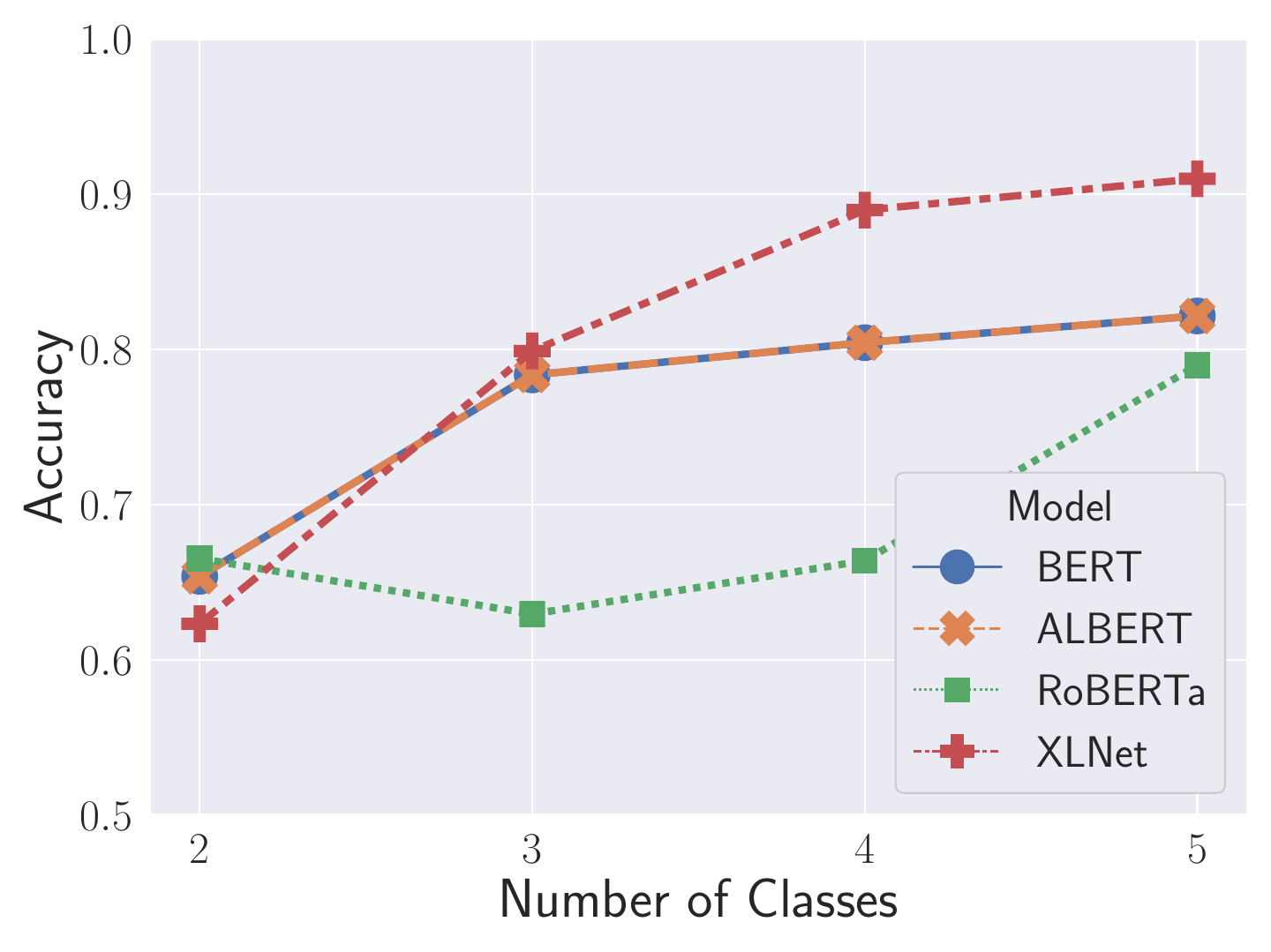}
\end{minipage}

}\hfill
\subfigure[Precision]{
\begin{minipage}[b]{0.32\textwidth}
    \includegraphics[width=1.0\textwidth]{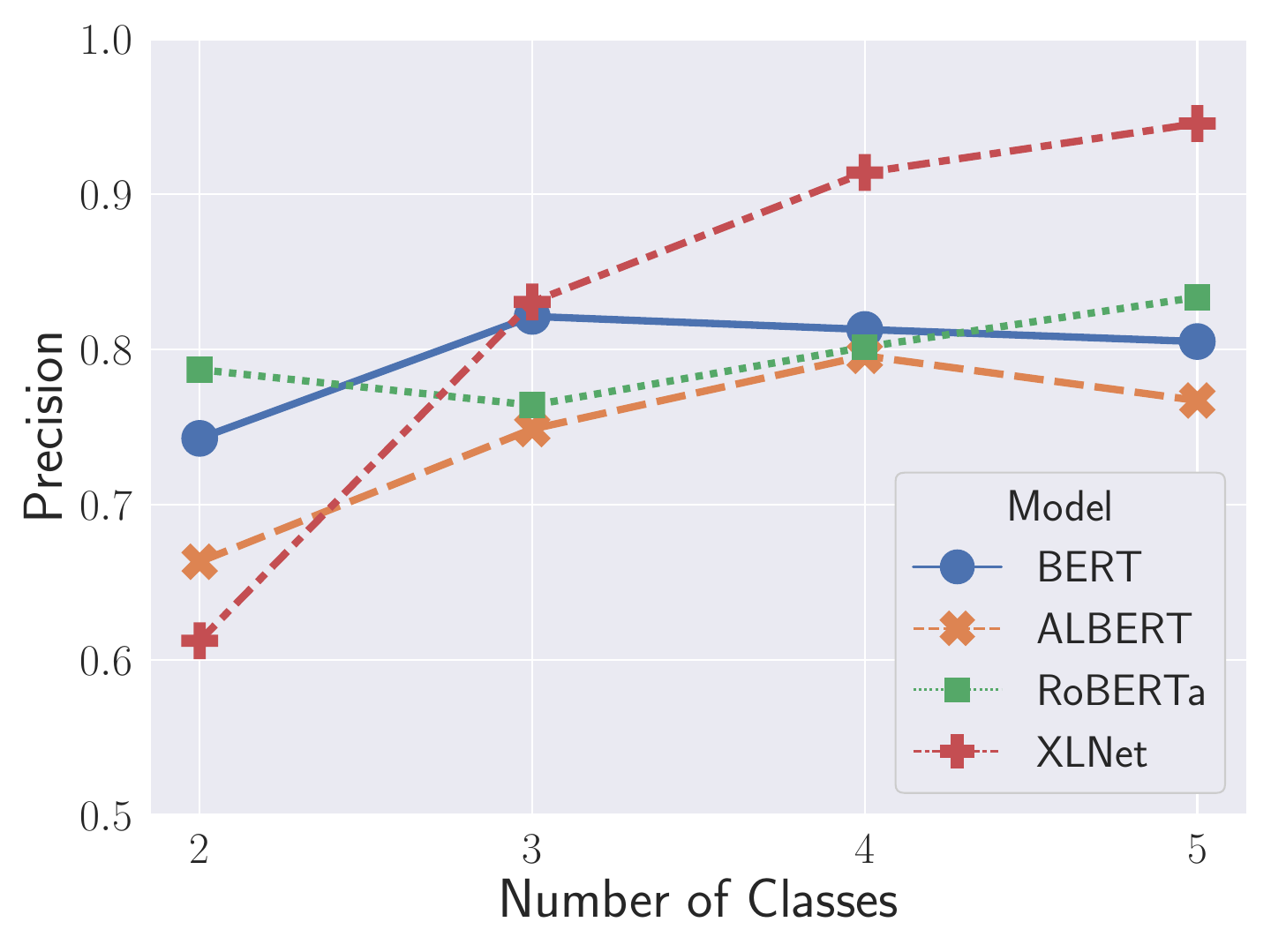}
\end{minipage}

}\hfill
\subfigure[Recall]{
\begin{minipage}[b]{0.32\textwidth}
    \includegraphics[width=1.0\textwidth]{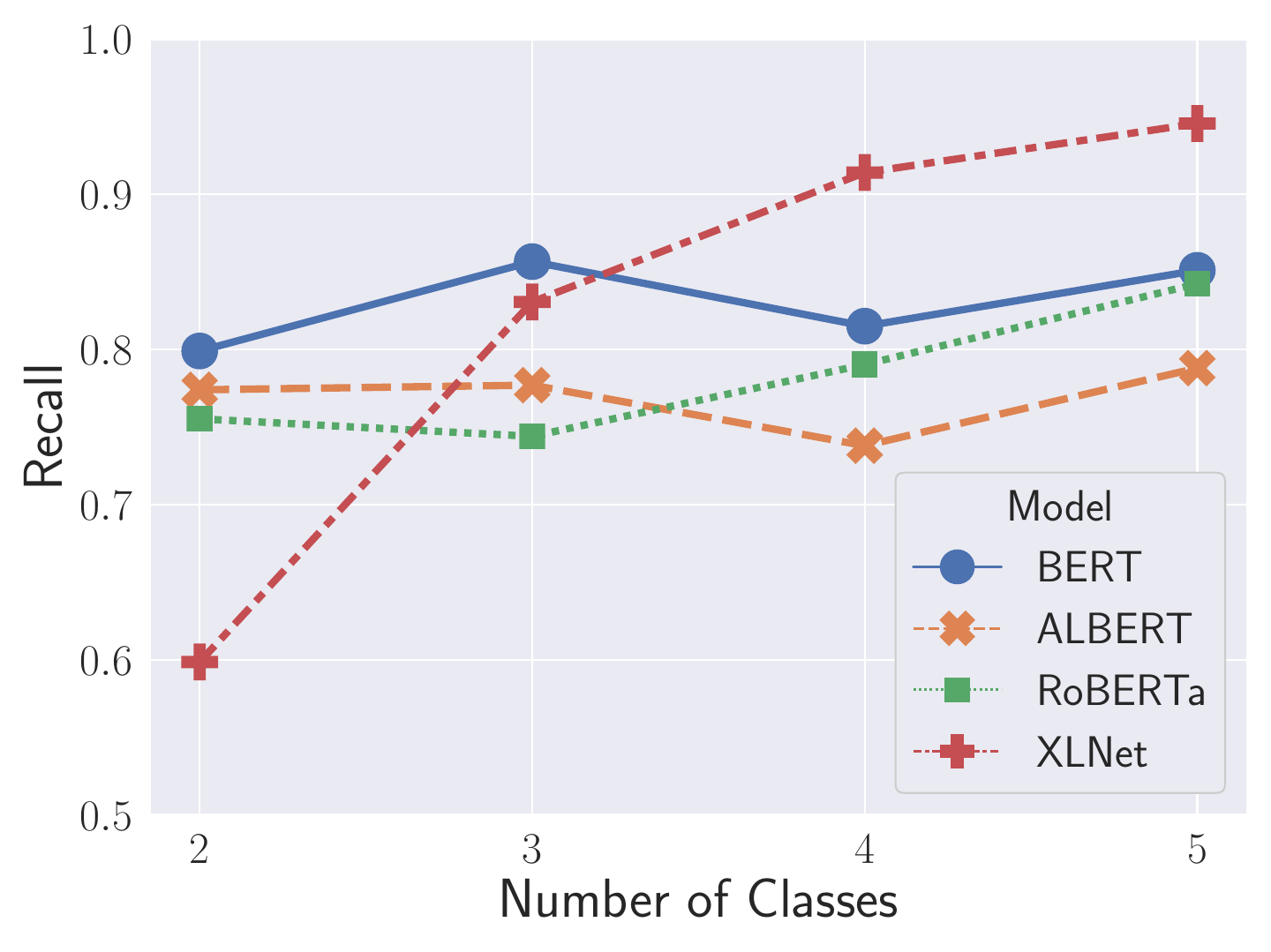}
\end{minipage}
}
\caption{Attack performance when varying the number of downstream task's label classes, conducted on Yelp Review Full dataset.}
\label{fig:num_class}
\vspace{10pt}
\end{figure*}

\section{Experiment Setup Details}
\subsection{Downstream Tasks \& Datasets}
\label{sec:appendix}
We list below the detailed description for the benchmark datasets adopted in this work and summarizes the key properties in Table~\ref{table:dataset}. We use the training split of the dataset for PLEs' finetuning unless otherwise specified.

\vspace{4pt}\noindent
\textbf{SST-2\footnote{\url{https://huggingface.co/datasets/sst2}}}: The Stanford Sentiment Treebank (SST) is a corpus with fully labeled parse trees that allows for a complete analysis of the compositional effects of sentiment in language. The corpus is based on the dataset introduced by~\cite{pang2005seeing} and consists of 11,855 single sentences extracted from movie reviews.  It was parsed using the Stanford parser, yielding a collection of 215,154 unique phrases derived from the parsed trees. The corpus is often used in binary classification experiments, where full sentences are categorized as either ``negative/somewhat negative'' or ``positive/somewhat positive'', with neutral sentences being discarded. This binary dataset is commonly referred to as SST-2 or SST binary. It encapsulates a total of 70,000 sentences, of which 67,300 are used for training.

\vspace{4pt}
\noindent\textbf{AG's News\footnote{\url{https://huggingface.co/datasets/ag_news}}}: The AG's News dataset serves as a benchmark dataset for topic classification tasks. It is an extensive collection of over a million news articles, amassed from more than 2,000 news sources by the academic news search engine, ComeToMyHead, over a year-long period. The dataset has been utilized as a text classification benchmark in research, notably in the study by \citet{ZZL15}. AG's News is structured as a four-class classification framework. In total, it encompasses 127,600 sentences, of which 120,000 from the training split are typically used for model finetuning. 

\vspace{4pt}
\noindent\textbf{Amazon Review}\footnote{\url{https://www.kaggle.com/datasets/kritanjalijain/amazon-reviews}}: The Amazon reviews dataset consists of reviews from amazon~\cite{ZZL15}. The data span a period of 18 years, including around 35 million reviews up to March 2013. Reviews include product and user information, ratings, and a plaintext review. The Amazon reviews polarity dataset is  is devised by designating reviews with scores 1 and 2 as negative (class 1) and those with scores 4 and 5 as positive (class 2). Reviews with a neutral score of 3 are excluded. The dataset is balanced with each class containing 1.8 million training samples and 20,000 testing samples.

\vspace{4pt}
\noindent\textbf{Yelp Review Full\footnote{\url{https://huggingface.co/datasets/yelp\_review\_full}}}: 
The Yelp reviews dataset is comprised of reviews collected from Yelp, originally extracted from the Yelp Dataset Challenge conducted in 2015. It was first introduced as a text classification benchmark by \citet{ZZL15}.  The dataset contains 650k trainig samples in total and 50k samples reserved for testing. 

\vspace{4pt}
\noindent\textbf{CoNLL2003\footnote{\url{https://huggingface.co/datasets/conll2003}}}: CoNLL2003 is used for named entity recognition (NER) %
adopting IOB2 tagging scheme, which includes 9 classes for each token: \{'O','B-PER','I-PER','B-ORG','I-ORG','B-LOC','I-LOC', 'B-MISC','I-MISC'\}.
It contains totally 20.7k sentences with the size of training set being 14k.

\vspace{4pt}
\noindent\textbf{SQuADv1.0\footnote{\url{https://huggingface.co/datasets/squad}}}: SQuADv1.0  is a reading comprehension dataset used for Question Answering(Q\&A)~\cite{2016arXiv160605250R}, which consists of questions posed by crowdworkers on a set of wikipedia articles, where the answer to every question is a segment of text, or span, from the corresponding reading passage, or the question might be unanswerable. 
The primary task involves providing a piece of context and a question, the answer to which is contained within the given context. The objective is to predict the ``start'' and ``end'' locations for the answer in the given context. This is typically modeled by predicting the start and end probabilities for each token within the given context. Subsequently, the answer is pinpointed within the two tokens that exhibit the highest predicted ``start'' and ``end'' probabilities. The dataset offers 87,600 sentences for training and an additional 10,600 sentences for testing.
\begin{table}[h]
\centering
\caption{Summary of different datasets.}
\vspace{8pt}
\label{table:dataset}
\resizebox{\columnwidth}{!}{
\begin{tabular}{cccc}
\toprule
\textbf{Dataset} & \textbf{\#Sentences} &   \textbf{Topic} & \textbf{Domain}\\
\midrule
SST-2 & 70k  & classification & movie review\\
AG's News &127.6k   & classification &  news article\\
Yelp Review & 650k & classification & reviews from Yelp\\
Amazon Review &  9100k  &  classification & amazon website review\\
CoNLL2003 &  20.7k  & NER & news wire articles \\
SQuADv1.0 & 98.2k   &    Q\&A& wikipedia articles  \\
\bottomrule
\end{tabular}
}

\end{table}

\section{Implementation Details}
The attack model is implemented as a three-layer fully connected network, i.e., 3-layer MLP, with ReLU activation functions after the first two layers and a final softmax layer to produce output probabilities. And we use the Adam optimizer with a learning rate 0.01 to train the attack model, following the setting of previous works~\cite{HLGZ22,CTCP21}. The input dimension of the attack model varies acorss different downstream tasks: 
\begin{itemize}[topsep=2pt,itemsep=2pt]
    \item For \textit{classification tasks} the dimension equals to number of label classes in the corresponding task: 2 for SST-2, 4 for AG's News, and 5 for both Yelp and Amazon Review dataset.
    \item The \textit{NER} is conducted as a token classification task where the classification output are 9 NER tags with IOB2 tagging scheme, so the input dimension for attack model is 9.
    \item For the \textit{Q\&A} task conducted on the SQuADv1.0 dataset, the fine-tuned downstream model will output two logits representing the ``start'' and ``end'' indicator, enabling the identification of answer positions. As a result, the input dimension for the attack model is twice the sequence length.
\end{itemize}

\section{Additional Results}

\paragraph{Effect of Downstream Task Configuration.}
We provide additional results on the relationship between attack performance and the configuration of downstream tasks, specifically, the number of classes in the classification task.
We conduct experiments on the Yelp Review Full dataset, which consists of five label classes in total. We then construct 4-class, 3-class, and 2-class subset by  dropping samples from one (randomly) selected label class, and we additionally fix the total number of training samples to be the same (i.e., 120,000), in order to eliminate the effect of dataset size.

As illustrated in  \figureautorefname~\ref{fig:num_class}, there is a positive correlation between attack performance and the number of classes. This suggests that an increase in the number of classes may lead to a higher amount of pre-training data information being leaked by the downstream model. A simple explanation for this is that the dimensionality of the downstream model's output grows in line with the number of classes. This provides more information regarding the task but also regarding the membership, consequently leading to an enhanced performance in detecting pre-training data's leakage.

\end{document}